\documentclass[journal]{IEEEtran}
\usepackage{amsmath,amsfonts}
\usepackage{algorithmic}
\usepackage{algorithm}
\usepackage{array}
\usepackage[caption=false,font=normalsize,labelfont=sf,textfont=sf]{subfig}
\usepackage{textcomp}
\usepackage{stfloats}
\usepackage{url}
\usepackage{verbatim}
\usepackage{graphicx}
\usepackage{cite}
\usepackage{multirow} 
\usepackage{makecell}
\usepackage[colorlinks=true,
            linkcolor=blue,
            citecolor=blue,
            urlcolor=blue,
            bookmarks=true]{hyperref}
\usepackage{bm}
\usepackage{booktabs}
\usepackage{mathtools}
\usepackage{lineno}
\usepackage{amsthm}
\usepackage{amssymb}

\newtheorem{definition}{\textit{Definition}}
\newtheorem{proposition}{\textit{Proposition}}
\begin{document}

\title{Physen-Noise2Noise: Physics-Guided Self-Supervised Defocus Deblurring with Bias Correction under Low-Light Conditions}

\author{Ziyan Huang, Lang Wu, Hongji Wang, Yifei Liu, Dongliang Tang, and Hongqiao Wang
\thanks{(Corresponding authors: Hongqiao Wang; Dongliang Tang.)}
\thanks{Ziyan Huang, Hongji Wang, Yifei Liu, and Hongqiao Wang are with the School of Mathematics and Statistics, Central South University, Changsha, 410083, People’s Republic of China (e-mail: 242111091@csu.edu.cn; 18707976263@163.com; 31250181@csu.edu.cn; Hongqiao.Wang@csu.edu.cn).}%
\thanks{Lang Wu and Dongliang Tang are with the Key Laboratory for Micro/Nano Optoelectronic Devices of Ministry of Education, Hunan Provincial Key Laboratory of Low-Dimensional Structural Physics and Devices, School of Physics and Electronics, Hunan University, Changsha 410082, People’s Republic of China (e-mail: langwu@hnu.edu.cn; dltang@hnu.edu.cn).}%
}

\markboth{IEEE TRANSACTIONS ON COMPUTATIONAL IMAGING,~Vol.~X, No.~Y, MONTH YEAR}%
{Huang \MakeLowercase{\textit{et al.}}: Physen-Noise2Noise: Physics-Guided Self-Supervised Defocus Deblurring with Bias Correction under Low-Light Conditions}

\IEEEpubid{0000--0000/00\$00.00~\copyright~2021 IEEE}

\maketitle

\begin{abstract}
Low-light, long-exposure defocus deblurring remains a challenging problem due to the simultaneous presence of severe blur and complex biased noise. Existing methods typically rely on simplified noise assumptions, which limits their effectiveness under realistic imaging conditions. In this work, we propose Physen-Noise2Noise, a self-supervised deblurring framework guided by the physical model of defocus imaging, which leverages noisy multi-frame observations without requiring clean reference images. Unlike conventional Noise2Noise-based approaches that assume zero-mean noise, we derive a frequency-domain constraint inherent to the defocus imaging process and incorporate it into the learning framework via a learnable noise bias parameter. In addition, a multi-frame noisy initialization strategy is introduced to suppress complex biased noise prior to deblurring, providing a more stable starting point for reconstruction. This formulation explicitly models biased noise and enables joint bias correction and high-frequency detail recovery during training. Furthermore, we develop a pretrain–finetune variant to enhance robustness and generalization under challenging noise conditions. Extensive experiments on both simulation and real-world datasets demonstrate that the proposed method consistently outperforms state-of-the-art self-supervised approaches for defocus deblurring in the presence of complex biased noise.
\end{abstract}

\begin{IEEEkeywords}
Complex biased noise, Defocus deblurring, Low-light imaging, Self-supervised learning.
\end{IEEEkeywords}

\section{Introduction}
\label{sec:introduction}
\IEEEPARstart{D}{efocus} blur is a common degradation in practical imaging systems and can be formulated as an inverse problem in computational imaging, where the observed image is modeled as the convolution of a latent sharp image with a defocus point spread function (PSF), often further corrupted by noise. This degradation arises from mismatches between the sensor plane and the focal plane due to autofocus errors, depth variations, or inherent optical limitations, leading to significant loss of high-frequency details and degraded visual quality.

In many real-world low-light imaging applications, such as surveillance monitoring \cite{ai2020extreme}, microscopy \cite{qu2024self}, and Industrial Vision \cite{tao2024legan}, defocus blur can coexist with complex biased noise. Under low-light conditions, the limited photon budget often necessitates long exposure times to achieve sufficient signal levels. These strategies inevitably introduce a mixture of noise sources \cite{healey2002radiometric, tian2000noise, konnik2014high, arko2022pyxel, el2005cmos, cao2023physics}, including shot noise, dark current noise, thermal noise, and fixed-pattern noise. 

The noise in low-light long-exposure imaging can be categorized according to its physical origin and statistical behavior. Photon shot noise follows Poisson photon-arrival statistics and is signal-dependent, while dark-current shot noise arises from thermally generated electrons and is mainly affected by exposure time, temperature, and sensor properties. In this process, the photoelectron expectation carries the latent irradiance information, whereas dark-current-related components introduce an additive bias that may remain approximately stable over a short acquisition period under fixed exposure, gain, and temperature settings~\cite{cao2023physics}. In addition, practical sensors often exhibit structured fixed-pattern components, such as pixel-wise dark-signal non-uniformity and row/column readout patterns, which further introduce spatially varying but temporally stable bias~\cite{arko2022pyxel, cao2023physics}. Readout noise originates from electronic circuits and is commonly approximated as signal-independent Gaussian noise~\cite{el2005cmos, cao2023physics}. Although photon shot noise and random readout noise are already pronounced under low-light conditions, the stable bias-related components are particularly difficult to compensate because their spatial distributions and statistical parameters depend on operating conditions such as exposure time, temperature, gain, and sensor characteristics.

Moreover, in-camera processing can introduce additional spatial correlations and alter sensor-noise statistics~\cite{chatterjee2011noise, jin2020review, lee2022ap}, further violating the independent and zero-mean noise assumptions commonly used in conventional deblurring models. Combined with defocus-induced high-frequency attenuation, signal-dependent noise, and stable sensor bias, this makes low-light long-exposure deblurring particularly ill-posed and motivates methods specifically designed for this setting.

\IEEEpubidadjcol

While numerous deblurring algorithms excel under well-lit conditions \cite{zhang2017beyond, liang2021swinir, chen2023masked, herbreteau2023normalization, li2024synthetic, dong2021learning, sanghvi2022photon, zhang2023infwide, dong2020deep, gong2020learning, Kageyama:20, Wei:20}, they falter in low-light environments. They struggle to distinguish genuine high-frequency edges from noise-induced perturbations, often resulting in either severe artifacts or over-smoothing. This limitation stems from insufficient exploration of the frequency-domain constraints inherent to defocus blur and the complex statistical properties of real-world noise. Recent self-supervised approaches like SSI \cite{kobayashi2020image} assume zero-mean, independent noise, failing with biased noise; UNID \cite{chen2022nonblind} lacks explicit noise modeling; and RLSN2N \cite{qu2024self}, despite combining Richardson-Lucy with Noise2Noise, is constrained by its handling of correlated noise and does not address bias. These observations reveal a fundamental challenge in defocus deblurring under low-light conditions: the mismatch between the deterministic forward degradation model and the complex, biased and spatially correlated noise encountered in real imaging systems. Existing methods typically rely on simplified assumptions such as zero-mean or independent noise, which are frequently violated in practice, leading to degraded restoration quality. As a result, it remains challenging to effectively integrate physical priors with realistic noise statistics in a self-supervised setting.

The random noise in adjacent frames is often approximately independent, while its bias component can remain stable over short acquisition sequences~\cite{wiedemann2024deep,wach2004noise}. This motivates multi-frame collaboration for initial denoising. Meanwhile, defocus blur attenuates high-frequency components, whereas observational noise perturbs these already weakened frequencies. Therefore, effective restoration should not rely on perfect pre-denoising alone, but should exploit the mismatch between PSF-induced high-frequency attenuation and noise-induced disturbance. This motivates our iterative bias-correction mechanism, which progressively compensates for stable noise bias and suppresses artifacts during self-supervised deblurring.

This work addresses defocus deblurring under low-light, long-exposure conditions. We propose a self-supervised deep learning framework, Physen-Noise2Noise (PN2N), which requires no clean reference images and incorporates the physical forward model of defocus blur into the optimization process while exploiting multi-frame observations for deblurring. Guided by the validated high-frequency attenuation constraint of defocus blur, the method first initializes the latent image using multiple noisy observations and then introduces a learnable bias parameter during reconstruction to adaptively correct biased noise. The main contributions of this work are summarized as follows:
\begin{itemize}
    \item We analyze the frequency-domain constraint of defocus blur, particularly high-frequency attenuation, and introduce a learnable bias to model biased noise, enabling adaptive correction of noise bias for improved restoration.
    
    \item We propose a self-supervised framework that requires no clean references, leveraging multi-frame observations and embedding the defocus forward model into optimization to achieve effective deblurring under low-light, long-exposure conditions.
    
    \item The proposed method delivers robust and high-quality deblurring under complex biased noise with minimal prior assumptions, maintaining stable performance in challenging real-world scenarios.
\end{itemize}

\section{Related Works}
\label{sec:Relatedworks}
\subsection{Nonblind image deblurring}
\label{subsec:NID}
WD~\cite{dhawan1985image}, LRA~\cite{lucy1974iterative, richardson1972bayesian}, and its extension NLR~\cite{mukherjee2018imaging} are classical nonblind deblurring methods. However, their simplified noise assumptions, such as known power spectra for WD and Poisson noise for LRA/NLR, often mismatch real blurred observations and cause artifacts or over-smoothing.

Unlike traditional approaches, deep learning–based methods have recently made remarkable progress in image deblurring. Several works \cite{hendriksen2020noise2inverse, ren2020neural, tang2023uncertainty, chen2022nonblind, qin2025robust} follow an autoregressive learning paradigm: the network output is convolved with the defocus PSF and compared to the observed blurred image to compute the loss, yielding promising results without paired training data. Other approaches also contribute to the area: deep image prior based methods \cite{ulyanov2018deep, wang2019image} exploit network architecture as an implicit regularizer, while techniques such as denoising the deblurring result \cite{nair2022nbd, qu2024self} offer alternative solutions. Nevertheless, these methods are generally outperformed by autoregressive approaches. Additionally, other research that relies on external resources, such as using CycleGAN for domain translation \cite{lim2020cyclegan}, is not fully self-supervised.

\noindent \textbf{Nonblind deblurring under noisy conditions.} Wiener deconvolution \cite{dhawan1985image}, the Richardson–Lucy algorithm  \cite{lucy1974iterative, richardson1972bayesian}, and nonlinear reconstruction \cite{mukherjee2018imaging} enhance robustness to noise by incorporating explicit noise models into the image formation process. In contrast, SSI \cite{kobayashi2020image}, inspired by Noise2Void \cite{krull2019noise2void}, adopts a mask-based learning strategy to achieve robustness; however, it relies on assumptions of spatially independent and unbiased noise. UNID \cite{chen2022nonblind} and \cite{qin2025robust} employ Monte Carlo Dropout \cite{gal2016dropout} to improve robustness under relatively mild noise conditions. Nevertheless, the lack of explicit noise modeling limits performance in more severe scenarios. RLSN2N \cite{qu2024self} combines the classical Richardson–Lucy algorithm \cite{lucy1974iterative, richardson1972bayesian} with the Noise2Noise paradigm \cite{lehtinen2018noise2noise}, decomposing the restoration process into separate denoising and deblurring stages. However, it does not exploit the frequency-domain relationship between the noise distribution and the blurring process.

\subsection{Image denoising}
\label{subsec:ID}
Traditional denoising methods include filtering~\cite{yang1996structure, takeda2007kernel}, transform-domain techniques~\cite{dabov2007image, zhang2005multiscale}, and total-variation-based methods~\cite{beck2009fast}, but they often suffer from high computational cost and loss of local details. Deep learning methods have recently become dominant due to their strong restoration performance~\cite{tian2020deep, izadi2023image, elad2023image}. Supervised denoisers~\cite{zhang2017beyond, liang2021swinir, chen2023masked, herbreteau2023normalization} achieve state-of-the-art results, but require large-scale paired clean/noisy training data, which is costly to collect in practice.

To alleviate this issue, a growing trend has emerged toward self-supervised denoising methods \cite{laine2019high}. A representative work is Noise2Noise (N2N) \cite{lehtinen2018noise2noise}, which recovers clean images from pairs of noisy images by learning to map one noisy observation to another under the assumption that the noise is zero-mean. However, N2N requires multiple noisy images, which limits its applicability. To overcome this limitation, one line of research, such as recorrupted-to-recorrupted \cite{pang2021recorrupted} and generalized recorrupted-to-recorrupted \cite{monroy2025generalized}, simulates training pairs by decomposing a single image but requires specifying the noise model and its parameters. ZS-DeconvNet \cite{qiao2024zero} inherits both the noise modeling strategy and its limitation, while further incorporating a deconvolution prior to achieve joint denoising and super-resolution. Another line, including Noise2Void \cite{krull2019noise2void}, Noise2Self \cite{huang2021neighbor2neighbor}, Neighbor2Neighbor \cite{huang2021neighbor2neighbor}, and Self2Self \cite{quan2020self2self}, leverages image spatial correlations to work with only a single image but introduces the additional assumption that noise is spatially independent. ZS-SIM \cite{11175047} and PRS-SIM \cite{chen2024self} propose more systematic pixel-realignment strategies within this paradigm, yet still rely on the same assumptions of unbiased and spatially independent noise. More recent works, such as Self-inspired Noise2Noise (SN2N) \cite{qu2024self} and Pixel2Pixel \cite{ma2025pixel2pixel}, relax the assumption of spatial independence while retaining the assumption of unbiased noise. DeepDeWedge \cite{wiedemann2024deep} constructs noise pairs from adjacent frames for training, thereby overcoming the data requirement of N2N, but does not account for the impact of biased noise.

\section{Methods}
\label{sec:methods}
\subsection{Degradation model}
\label{subsec:Degradation_model}
In defocus blur, light rays from a point source do not converge to a single point but instead form a disk, known as the circle of confusion, rather than a single point on the sensor. The point spread function (PSF) characterizes the impulse response of the imaging system to a point source. In many practical scenarios, particularly under relatively small aperture settings where the depth of field is large, the defocus blur can be reasonably approximated as spatially invariant. This assumption mainly applies to controlled or locally uniform imaging settings, where the object depth variation is limited within the considered field of view and the optical configuration remains fixed during acquisition. For scenes with strong depth variation, wide-field aberrations, or spatially varying defocus, the proposed model can be applied locally with patch-wise PSFs, but direct handling of fully spatially varying blur is beyond the scope of this work. Under this adopted assumption, the degraded imaging process is formulated as a convolution between the sharp image $x$ and the PSF:

\begin{equation}
  y = PSF \otimes x + n,
  \label{eq:1}
\end{equation}
where $y$ denotes the blurred image, $\otimes$ represents the convolution operator, $PSF \otimes x$ models the process of light propagation, and $n$ accounts for the additive noise, which typically exhibits a complex biased distribution \cite{healey2002radiometric, tian2000noise, konnik2014high}. This reconstruction is formulated as an optimization problem:
\begin{equation}
\hat{x} = \arg\min_x \| PSF \otimes x - y \|_2^2.
  \label{eq:2}
\end{equation}

This ill-posed inverse problem presents significant challenges: even minor noise can lead to substantial artifacts in the reconstruction due to the inherent instability of the solution. These challenges underscore the need for defocus deblurring algorithms operable in complex biased noise distributions.

\subsection{Frequency domain constraint and biased noise residual cue}
\label{subsec:constraint}

For the noise-free part of the degradation model in Eq.~\eqref{eq:1}, we write \(y=p\otimes x\), where \(p\) denotes the PSF. In the Fourier domain, this model imposes a feasibility constraint: the spectrum of \(y\) must be compatible with the PSF frequency response. We use a continuous-domain formulation to characterize this constraint and then connect it to the discrete implementation. The full proof is provided in the supplementary material.

Let \(\mathcal I=\mathcal O=L^2(\mathbb R^2)\), \(p\in L^1(\mathbb R^2)\cap L^2(\mathbb R^2)\), and define
\begin{equation}
    T_p x=p\otimes x .
  \label{eq:3}
\end{equation}

Let
\begin{equation}
    P(\omega)=\mathcal F\{p\}(\omega), \omega=(u,v)\in\Omega=\mathbb R^2.
  \label{eq:4}
\end{equation}
be the PSF frequency response. By Plancherel's and convolution theorem, a noise-free blurred observation \(y\) satisfies
\begin{equation}
    \widehat y(\omega)=P(\omega)\widehat x(\omega),
  \label{eq:5}
\end{equation}
where \(\widehat x\) and \(\widehat y\) are the Fourier transforms of \(x\) and \(y\). According to Fourier optics theory~\cite{goodman2005introduction} and previous works~\cite{chen2022nonblind, hosseini2019convolutional, boreman2001modulation}, defocus PSFs attenuate high-frequency components and may contain null or near-null frequency regions. We formalize the PSFs considered in this work as follows.

\begin{definition}[PSF class considered in this work]
\label{def:PSF_frequency}
Let \(P(\omega)=\mathcal F\{p\}(\omega)\). The PSF considered in this work satisfies one of the following conditions:
\begin{enumerate}
    \item \label{cond:1} The zero set \(Z_p=\{\omega\in\Omega:P(\omega)=0\}\) has positive Lebesgue measure.
    \item \label{cond:2} \(P(\omega)\neq0\) almost everywhere and \(\lim_{|\omega|\to\infty}|P(\omega)|=0\).
\end{enumerate}
\end{definition}

Condition~(\ref{cond:1}) covers PSFs with null frequency regions, and Condition~(\ref{cond:2}) covers nonzero but high-frequency-decaying PSFs. The blurred observation space induced by the PSF is
\begin{equation}
    \mathcal V_p=\operatorname{Ran}(T_p)=\{y\in L^2(\mathbb R^2):\exists x\in L^2(\mathbb R^2), \ y=T_px\}.
  \label{eq:6}
\end{equation}

\begin{proposition}[Range characterization of PSF convolution]
\label{pro:source_condition}
Let \(T_p\) be the PSF convolution operator defined above. Then
\begin{equation}
\begin{split}
    \mathcal V_p = \Bigl\{ y\in L^2(\mathbb R^2):\ &
    \widehat y(\omega)=0 \text{ a.e. on } Z_p,\\
    &
    \int_{\Omega\setminus Z_p}
    \frac{|\widehat y(\omega)|^2}{|P(\omega)|^2}\,d\omega<\infty
    \Bigr\}.
\end{split}
\end{equation}
Moreover, if the PSF satisfies either Condition~(\ref{cond:1}) or Condition~(\ref{cond:2}), then
\begin{equation}
    \mathcal V_p\subsetneq\mathcal I .
  \label{eq:8}
\end{equation}
\end{proposition}

Proposition~\ref{pro:source_condition} shows that a physically feasible blurred observation cannot contain arbitrary frequency-domain variations. Specifically, frequency components in the null region of the PSF must vanish almost everywhere, and components in highly attenuated frequency regions must be sufficiently weak such that $\widehat y/P$ remains square-integrable. Therefore, the PSF does not merely blur an image visually; it also defines a source condition that every noise-free observation must satisfy.

\begin{figure}[ht]
\centering\includegraphics[width=3.5in]{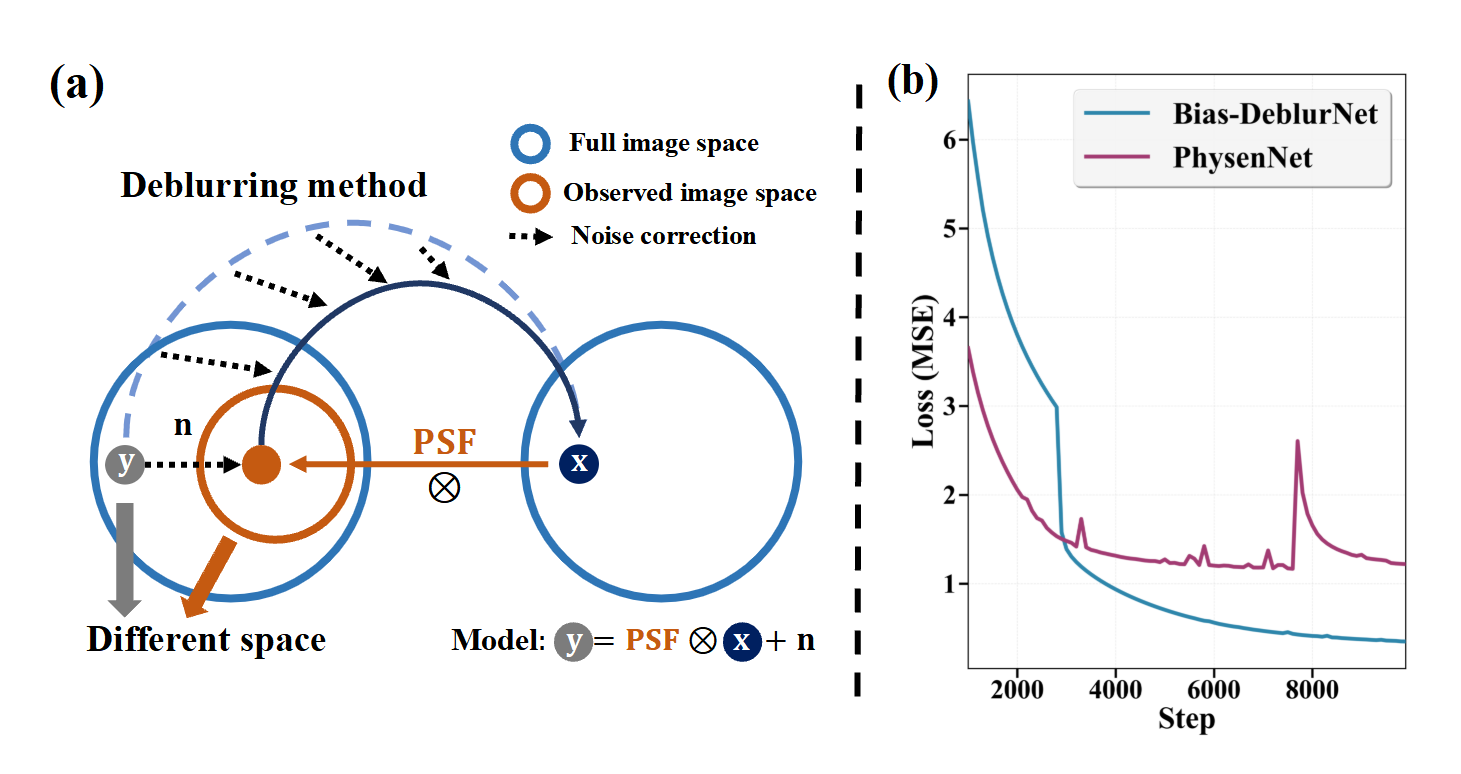}
  \caption{(a) Impact of the defocus blur frequency constraint on deblurring. (b) A representative example of loss descent curves, comparing our joint optimization of deblurring and noise bias removal against conventional deblurring-only optimization. PhysenNet \cite{wang2020phase} is a method based on Eq. (\ref{eq:2}).}
  \label{fig:theory}
\end{figure}

\noindent \textbf{Connection to the discrete implementation.}
The continuous source condition has a direct finite-dimensional analogue. Under circular boundary conditions, or the standard block-circulant approximation of convolution, the discrete PSF operator is diagonalized by the unitary DFT:
\begin{equation}
    H=F^\ast\operatorname{diag}(P_1,\ldots,P_N)F,
    \qquad
    \widehat y_k=P_k\widehat x_k .
  \label{eq:9}
\end{equation}
Thus, zero PSF responses impose hard constraints on feasible observations, while small PSF responses impose stability constraints. Specifically, for any energy-bounded object image \(\|x\|_2\leq E\), a feasible blurred observation \(y=Hx\) satisfies
\begin{equation}
    \sum_{k:P_k\neq0}
    \frac{|\widehat y_k|^2}{|P_k|^2}
    =
    \sum_{k:P_k\neq0}
    |\widehat x_k|^2
    \leq
    E^2 .
    \label{eq:discrete_ellipsoid}
\end{equation}
Therefore, frequencies with small \(|P_k|\) can contain only limited energy; otherwise, they would require unrealistically large latent-image energy. We define the PSF-attenuated frequency set as
\begin{equation}
    \mathcal K_\tau=\{k:|P_k|<\tau\},
\end{equation}
and let \(Q_\tau=F^\ast M_\tau F\) be the corresponding DFT-domain projection, where \(M_\tau\) keeps the coefficients in \(\mathcal K_\tau\) and zeros out the rest. Then, for any clean blurred observation \(y=Hx\) with \(\|x\|_2\leq E\),
\begin{equation}
    \|Q_\tau y\|_2
    =
    \|Q_\tau Hx\|_2
    \leq
    \tau E .
    \label{eq:discrete_attenuation}
\end{equation}
Eq.~\eqref{eq:discrete_attenuation} is the finite-dimensional counterpart of Proposition~\ref{pro:source_condition}.

\noindent \textbf{Biased noise and residual cue.}
We further explain why biased noise can leave a learnable cue in the residual of a blur-consistency model. Consider the discrete observation model
\begin{equation}
    z_i=Hx_i+b_i+n_i,
  \label{eq:13}
\end{equation}
where \(H\) is the discrete PSF convolution matrix, \(x_i\) is the latent clean image, \(b_i\) denotes biased noise, and \(n_i\) denotes zero-mean random noise. Let the deblurring network predict \(\hat x_i\), and define the blur-consistency residual as
\begin{equation}
    r_i
    =
    H\hat x_i-z_i
    =
    H(\hat x_i-x_i)-b_i-n_i .
    \label{eq:residual_def}
\end{equation}

Projecting onto \(\mathcal K_\tau\) gives
\begin{equation}
    Q_\tau r_i
    =
    Q_\tau H(\hat x_i-x_i)
    -
    Q_\tau b_i
    -
    Q_\tau n_i .
    \label{eq:projected_residual}
\end{equation}

The first term in Eq.~\eqref{eq:projected_residual} is the blur-explainable component. By the same attenuation argument as Eq.~\eqref{eq:discrete_attenuation}, it satisfies
\begin{equation}
    \|Q_\tau H(\hat x_i-x_i)\|_2
    \leq
    \tau\|Q_\tau(\hat x_i-x_i)\|_2 .
    \label{eq:residual_attenuation}
\end{equation}

Assume that the projected random noise is zero-mean and that the biased noise has a stable projected component:
\begin{equation}
    \mathbb E[Q_\tau n_i]=0,
    \mathbb E[Q_\tau b_i]=\mu_b,
    \mu_b\neq0 .
  \label{eq:17}
\end{equation}
Taking expectation in Eq.~\eqref{eq:projected_residual} and using Eq.~\eqref{eq:residual_attenuation} gives the following conservative lower bound. The detailed proof is provided in the supplementary material.

\begin{proposition}[Expected residual lower bound]
\label{pro:residual_fingerprint}
Under the above assumptions, the expected projected residual satisfies
\begin{equation}
    \|\mathbb E[Q_\tau r_i]\|_2
    \geq
    \left[
    \|\mu_b\|_2
    -
    \tau
    \mathbb E\|Q_\tau(\hat x_i-x_i)\|_2
    \right]_+ ,
    \label{eq:residual_fingerprint_lower_bound}
\end{equation}
where \([a]_+=\max(a,0)\).
\end{proposition}

Proposition~\ref{pro:residual_fingerprint} shows that the expected projected residual remains non-vanishing when the statistically stable biased component dominates the attenuated reconstruction error. Hence, under the stated stability and reconstruction-error conditions, a blur-only fitting process is not expected to suppress this component without leaving a persistent residual mismatch.

This residual cue also aligns with an energy-bounded or optimization perspective. A blur-only model can explain a residual component only by modifying the latent image before applying the PSF. For components that violate the PSF-induced source condition, such an explanation would require the latent image to compensate for weak PSF responses, leading to disproportionately large updates in PSF-attenuated frequency regions. These updates are disfavored by the energy-bounded reconstruction assumption, the implicit natural image prior \cite{ulyanov2018deep}, and the finite optimization dynamics of CNNs.

Therefore, introducing a shared learnable bias $b$ through
\begin{equation}
    z_i\approx H\hat x_i+\hat b ,
  \label{eq:19}
\end{equation}
the model obtains a lower-cost way to explain the PSF-consistency mismatch. The bias term is not multiplied by \(H\), and therefore it can model the statistically stable residual component, while the deblurring focuses on reconstructing structures that are compatible with the forward model. Moreover, since \(b\) is shared across samples, it acts as a sequence-level compensation: sample-specific fluctuations cannot be consistently explained by a single parameter and are averaged out in the aggregate objective, whereas statistically stable bias reduce the loss across samples and are preferentially captured.

This analysis should be interpreted as a conservative sufficient-condition argument rather than an exhaustive identifiability theorem. It clarifies why, in the low-light long-exposure setting with stable bias statistics, the blur-consistency residual can serve as a learnable cue for bias correction. As illustrated in Fig.~\ref{fig:theory}(a), Eq.~\eqref{eq:residual_fingerprint_lower_bound} indicates that a blur-only fitting process may retain a non-vanishing projected residual when the stable biased component dominates the attenuated reconstruction error. Introducing bias correction offers a lower-cost explanation for this residual, thereby guiding the optimization toward simultaneously achieving physically plausible deblurring and noise-bias compensation, as illustrated in Fig.~\ref{fig:theory}(b).

\subsection{Multi-frame initial denoising}
The proposed PN2N framework uses a two-step approach. It first leverages multi-frame noisy images for initial denoising and subsequently introduces a bias-adaptive physics-guided deep learning method to achieve noise bias compensation and defocus deblurring. Specifically, the Self-inspired Noise2Noise (SN2N) method \cite{qu2024self} is adopted for initialization.  

\begin{figure}[ht]
\centering\includegraphics[width=\linewidth]{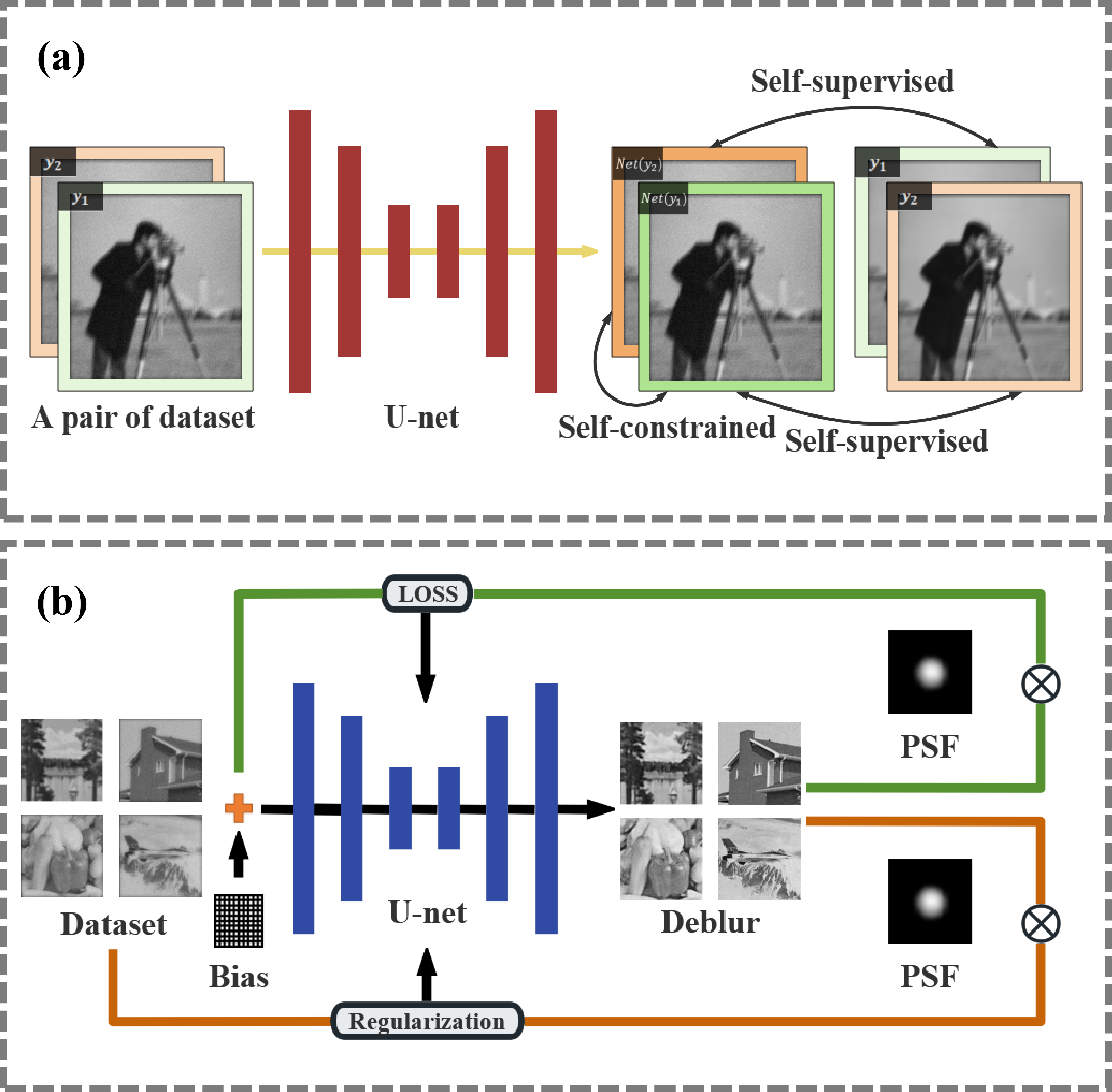}
  \caption{(a) The SN2N strategy: integration of noise-to-noise loss and self-constrained loss. (b) Bias-DeblurNet architecture: design of its loss function and regularization. The original images used in this figure are from \cite{deng2009imagenet, CVGUGR}.}
  \label{fig:method}
\end{figure}

Noise2Noise (N2N) \cite{lehtinen2018noise2noise} restores images using only pairs of noisy observations without requiring clean references. It relies on the assumption that the noise in different observations of the same signal is independent and identically distributed (\emph{i.i.d.}) with zero mean. Under this assumption, minimizing the mean squared error between noisy image pairs drives the network to estimate the expectation of the underlying clean signal. The SN2N further improves robustness by enforcing prediction consistency between different noisy observations, as shown in Fig. \ref{fig:method}(a). Given two noisy images $(y_1, y_2)$ from the same latent signal, the network $f_\theta$ is trained with

\begin{equation}
\begin{split}
\mathcal{L} = 
\mathbb{E}_{(y_1, y_2)} \Big[ 
|f_\theta(y_1) &- y_2|_2^2 + |f_\theta(y_2) - y_1|_2^2 \\
&+ \lambda_1 |f_\theta(y_1) - f_\theta(y_2)|_2^2 
\Big],
\label{eq:20}
\end{split}
\end{equation}
where $\lambda_1$ is a weighting coefficient. The first two terms follow the N2N, while the consistency term encourages identical predictions for different noisy observations of the same signal.

By splitting the multi-frame sequence into odd/even subsets, we can construct paired training samples \cite{wiedemann2024deep}. When the noise distribution is unbiased, the SN2N initialization enables a direct mapping from noisy blurred observations to the corresponding clean observations. However, under low-light and long-exposure conditions, the noise distribution becomes biased, the observed blurred images will be mapped to
\begin{equation}
y^{\prime} = PSF \otimes x + \mu,
  \label{eq:21}
\end{equation}
where $y'$ represents the denoised image obtained from $y$ via SN2N, and $\mu$ denotes the expectation of the noise distribution to which $n$ belongs, irregularly disrupting the distribution in $y^{\prime}$, which causes $y^{\prime}$ to lie outside the observed image space as shown in Fig. \ref{fig:theory}(a). When solving for $\hat{x}$ on such an image $y^{\prime}$, the result $PSF \otimes \hat{x}$ lies in the observed image space. Since $y^{\prime}$ itself still lies outside the space, there exists no $\hat{x}$ that can reduce the loss $| PSF \otimes \hat{x} - y^{\prime} |_2^2$ to zero, thus leading to optimization stagnation, as referred to in Section \ref{subsec:constraint}.

This observation reveals a fundamental limitation of existing self-supervised denoising-based pipelines: the bias in the denoised result violates the physical consistency required by the forward model, preventing effective optimization in subsequent deblurring. Notably, this inconsistency manifests as optimization stagnation, which provides a cue for estimating the underlying $\mu$. This insight motivates a joint formulation that explicitly accounts for noise bias during reconstruction.

\subsection{Bias-adaptive physics-guided deblurring}
We propose a bias-adaptive physics-guided deblurring network as step 2, termed Bias-DeblurNet, as illustrated in Fig.~\ref{fig:method}(b). It introduces a learnable parameter $b$ to compensate for the noise bias and jointly optimize it with the network:

\begin{equation}
\begin{split}
R_\theta, b = \arg\min_{\theta, b} \mathbb{E}_{y^{\prime}} \Big[ 
&\| PSF \otimes R_\theta(y^{\prime} - b) - (y^{\prime} - b) \|_2^2 \\
&+ \lambda_2 \| PSF \otimes R_\theta(y^{\prime} - b) - y^{\prime} \|_2^2 
\Big],
\end{split}
\label{eq:22}
\end{equation}
where $R_\theta$ denotes the reconstruction network with parameters $\theta$, $b$ denotes a learnable bias-compensation term, and $\lambda_2$ is the regularization coefficient. Here, $b$ is introduced to compensate for the stable bias component remaining in the SN2N-initialized observation, rather than to recover the entire noise realization. Under the same imaging system and fixed exposure/gain settings, different scenes may share stable sensor-dependent bias components, such as dark-current expectation, dark-signal non-uniformity, high-current tail or hot-pixel bias, and row/column fixed-pattern components. Therefore, $b$ is shared across a small number of observations selected from consecutive-frame sequences of different scenes or targets.

Although $b$ is formally a free parameter and may degenerate in single-sample optimization, such degeneration is discouraged in the multi-sample setting. A single shared bias term cannot consistently absorb scene-dependent structures from multiple different samples; fitting the content of one sample would increase the reconstruction or blur-consistency error for other samples. Therefore, the learned $b$ is encouraged to capture the scene-invariant bias component under the current acquisition condition rather than arbitrary scene content, and serves as a compensation term for improving the subsequent deblurring optimization. The second term in Eq.~(\ref{eq:22}) further acts as a bias regularizer. As shown in Fig.~\ref{fig:theory}(a), even an accurately restored sharp image, when convolved with the PSF, should remain consistent with the observed blurred image. Motivated by this observation, the regularization constrains the reblurred reconstruction $PSF\otimes R_\theta(y'-b)$ not to deviate excessively from the original observation $y'$. This prevents the learned $b$ from absorbing scene-dependent structures or PSF-mismatch residuals. Consequently, $b$ should be interpreted as a constrained bias-compensation term rather than an unconstrained pixel-wise estimator of the noise expectation. In the experiments, we evaluate this compensation by measuring the similarity between the corrected observation $y'-b$ and the clean blurred image $PSF\otimes x$.

To further reduce the data requirement, we propose pretrain-finetune Physen-Noise2Noise (PF-PN2N), a pretrain--finetune framework. A model is first pretrained on images with similar noise characteristics and then finetuned using a small amount of observations. This strategy enables efficient adaptation to new data distributions, achieving performance comparable to training from scratch with substantially larger datasets.

\section{Experiment}
\label{sec:experiment}
\subsection{Settings and details}
\label{subsec:Settings and Detail}

\noindent \textbf{Implementation details.}
The SN2N network follows the implementation in~\cite{lehtinen2018noise2noise, qu2024self}. Bias-DeblurNet is implemented as a U-Net network. Before feature extraction, the input is corrected by subtracting a learnable spatial bias term \(b\in\mathbb R^{1\times W\times H\times1}\), which is replicated across the batch dimension and optimized jointly with the network. Detailed network architectures, implementation environment, and hyperparameter analysis are provided in the supplementary material.

\noindent \textbf{Methods included for comparison.}
We compare PN2N with representative baselines from different deblurring and restoration paradigms. Classical non-learning deconvolution methods include Wiener deconvolution (WD)~\cite{dhawan1985image}, Richardson--Lucy algorithm (LRA)~\cite{lucy1974iterative, richardson1972bayesian}, and nonlinear reconstruction (NLR)~\cite{mukherjee2018imaging}. Recent self-supervised or unsupervised deblurring methods include SSI~\cite{kobayashi2020image}, UNID~\cite{chen2022nonblind}, and RL-SN2N~\cite{qu2024self}. We further include several restoration baselines from different technical routes: DPIR \cite{zhang2021plug} as a deep denoising-prior method, WDIP \cite{bredell2023wiener} as a blind deconvolution baseline, DiffPIR \cite{zhu2023denoising} as a diffusion-prior method, and DWDN \cite{dong2020deep} as a supervised baseline. These baselines cover classical deconvolution, self-supervised restoration, deep and diffusion priors, blind deconvolution, and supervised learning. Methods unsuitable for noisy degradation are excluded, and all baseline hyperparameters are tuned for best performance in each setting.

\begin{table*}[t]
\centering
\begin{minipage}{0.48\textwidth}
\centering
\caption{Ablation study results on simulation dataset.}
\label{tab:ablation1}
\footnotesize
\setlength{\tabcolsep}{1pt} 
\renewcommand{\arraystretch}{1.2}
\newcommand{\pmstd}[2]{\makecell[c]{#1\\[-1pt]{\scriptsize$(\pm #2)$}}}
\newcommand{\bpmstd}[2]{\makecell[c]{\textbf{#1}\\[-1pt]{\scriptsize\textbf{$(\pm #2)$}}}}
\begin{tabular}{@{}c c *{8}{c} @{}}
\toprule
& & \multicolumn{2}{c}{T1} & \multicolumn{2}{c}{T2} & \multicolumn{2}{c}{T3} & \multicolumn{2}{c}{T4} \\
& PSF & PSNR$\uparrow$ & SSIM$\uparrow$ & PSNR$\uparrow$ & SSIM$\uparrow$ & PSNR$\uparrow$ & SSIM$\uparrow$ & PSNR$\uparrow$ & SSIM$\uparrow$\\
\midrule

& 1 & \pmstd{26.75}{0.09} & \pmstd{0.9258}{0.001} & \pmstd{15.83}{0.56} & \pmstd{0.2920}{0.016} & \pmstd{27.08}{1.38} & \pmstd{0.8683}{0.068} & \bpmstd{31.29}{0.41} & \bpmstd{0.9306}{0.004} \\
& 2 & \pmstd{23.61}{0.07} & \pmstd{0.8638}{0.002} & \pmstd{17.00}{0.16} & \pmstd{0.3956}{0.003} & \pmstd{27.84}{1.73} & \pmstd{0.8561}{0.115} & \bpmstd{30.14}{0.39} & \bpmstd{0.9029}{0.010} \\
& 3 & \pmstd{24.81}{0.03} & \pmstd{0.8548}{0.001} & \pmstd{18.31}{0.99} & \pmstd{0.3667}{0.040} & \pmstd{25.42}{0.46} & \pmstd{0.7775}{0.023} & \bpmstd{28.08}{0.18} & \bpmstd{0.8617}{0.014} \\
& 4 & \pmstd{22.11}{0.09} & \pmstd{0.7301}{0.002} & \pmstd{16.77}{0.33} & \pmstd{0.3418}{0.014} & \pmstd{23.07}{0.22} & \pmstd{0.7136}{0.018} & \bpmstd{26.35}{0.47} & \bpmstd{0.7872}{0.017} \\
& 5 & \pmstd{18.60}{0.14} & \pmstd{0.5054}{0.007} & \pmstd{17.05}{0.28} & \pmstd{0.4015}{0.014} & \pmstd{21.07}{0.32} & \pmstd{0.5188}{0.021} & \bpmstd{23.40}{1.36} & \bpmstd{0.6239}{0.030} \\ 
\bottomrule
\end{tabular}
\end{minipage}
\begin{minipage}{0.48\textwidth}
\centering
\caption{Model configuration. $\checkmark$ indicates that the module is included.}
\label{tab:ablation2}
\footnotesize
\setlength{\tabcolsep}{2pt} 
\renewcommand{\arraystretch}{1.2}
\begin{tabular}{@{}c c *{2}{c} @{}}
\toprule
Configuration & SN2N initialization  & Bias-adaptive & DeblurNet \\
\midrule
T1 & $\checkmark$ & $\times$  & $\times$  \\
T2 & $\times$  & $\checkmark$ & $\checkmark$ \\
T3 & $\checkmark$ & $\times$  & $\checkmark$ \\
T4 & $\checkmark$ & $\checkmark$ & $\checkmark$ \\
\bottomrule
\end{tabular}
\end{minipage}
\end{table*}

\begin{table*}[htbp]
\centering
\scriptsize
\caption{Performance comparison of different methods on controlled biased noise simulation dataset}
\label{tab:simulation_compare}
\setlength{\tabcolsep}{0.75pt}
\renewcommand{\arraystretch}{1}
\begin{tabular}{@{}c c *{22}{c} @{}} 
\toprule
& & \multicolumn{2}{c}{NLR} & \multicolumn{2}{c}{LRA} & \multicolumn{2}{c}{WD} & \multicolumn{2}{c}{SSI} & \multicolumn{2}{c}{UNID} & \multicolumn{2}{c}{RL-SN2N} & \multicolumn{2}{c}{PN2N} & \multicolumn{2}{c}{DPIR} & \multicolumn{2}{c}{WDIP} & \multicolumn{2}{c}{DiffPIR} & \multicolumn{2}{c}{DWDN}\\
& PSF & PSNR$\uparrow$ & SSIM$\uparrow$ & PSNR$\uparrow$ & SSIM$\uparrow$ & PSNR$\uparrow$ & SSIM$\uparrow$ & PSNR$\uparrow$ & SSIM$\uparrow$ & PSNR$\uparrow$ & SSIM$\uparrow$ & PSNR$\uparrow$ & SSIM$\uparrow$ & PSNR$\uparrow$ & SSIM$\uparrow$ & PSNR$\uparrow$ & SSIM$\uparrow$ & PSNR$\uparrow$ & SSIM$\uparrow$ & PSNR$\uparrow$ & SSIM$\uparrow$ & PSNR$\uparrow$ & SSIM$\uparrow$\\
\midrule
& 1 & 15.51 & 0.6560 & 23.17 & 0.7243 & 28.62 & 0.8590 & 23.22 & 0.7464 & 27.58 & 0.8901 & 26.77 & 0.7988 & \textbf{31.29} & \textbf{0.9306} & 25.87 & 0.8146 & 24.84 & 0.6281 & 24.87 & 0.7601 & 25.35 & 0.7270\\
& 2 & 19.47 & 0.7154 & 23.10 & 0.7521 & 26.34 & 0.8333 & 25.38 & 0.8930 & 23.44 & 0.8485 & 24.31 & 0.7439 & \textbf{30.14} & \textbf{0.9029} & 24.70 & 0.7732 & 20.69 & 0.4572 & 23.14 & 0.6994 & 25.25 & 0.6917\\
& 3 & 16.82  & 0.6725  & 24.00  & 0.7788 & 26.29 & 0.8435 & 25.74 & 0.8553 & 24.84 & 0.8609 & 23.28 & 0.7850 & \textbf{28.08} & \textbf{0.8617} & 24.72 & 0.7678 & 21.78 & 0.4688 & 22.96 & 0.7011 & 26.19 & 0.7969\\
& 4 & 17.88  & 0.6261 & 20.82 & 0.6780 & 23.40 & 0.7463 & 22.07 & 0.6711 & 21.89 & 0.7170 & 19.59 & 0.6725 & \textbf{26.35} & \textbf{0.7872} & 23.44 & 0.6906 & 18.42 & 0.2556 & 21.13 & 0.6155 & 24.80 & 0.7643\\
& 5 & 18.42 & 0.5054 & 15.05 & 0.3828 & 18.61 & 0.5142 & 17.73 & 0.4012 & 17.55 & 0.4376 & 15.46 & 0.4323 & \textbf{23.40} & \textbf{0.6239} & 18.22 & 0.4789 & 13.42 & 0.0921 & 16.51 & 0.3742 & 20.34 & 0.5696\\
\bottomrule
\end{tabular}
\end{table*}

\begin{figure*}[ht]
\centering\includegraphics[width=7in]{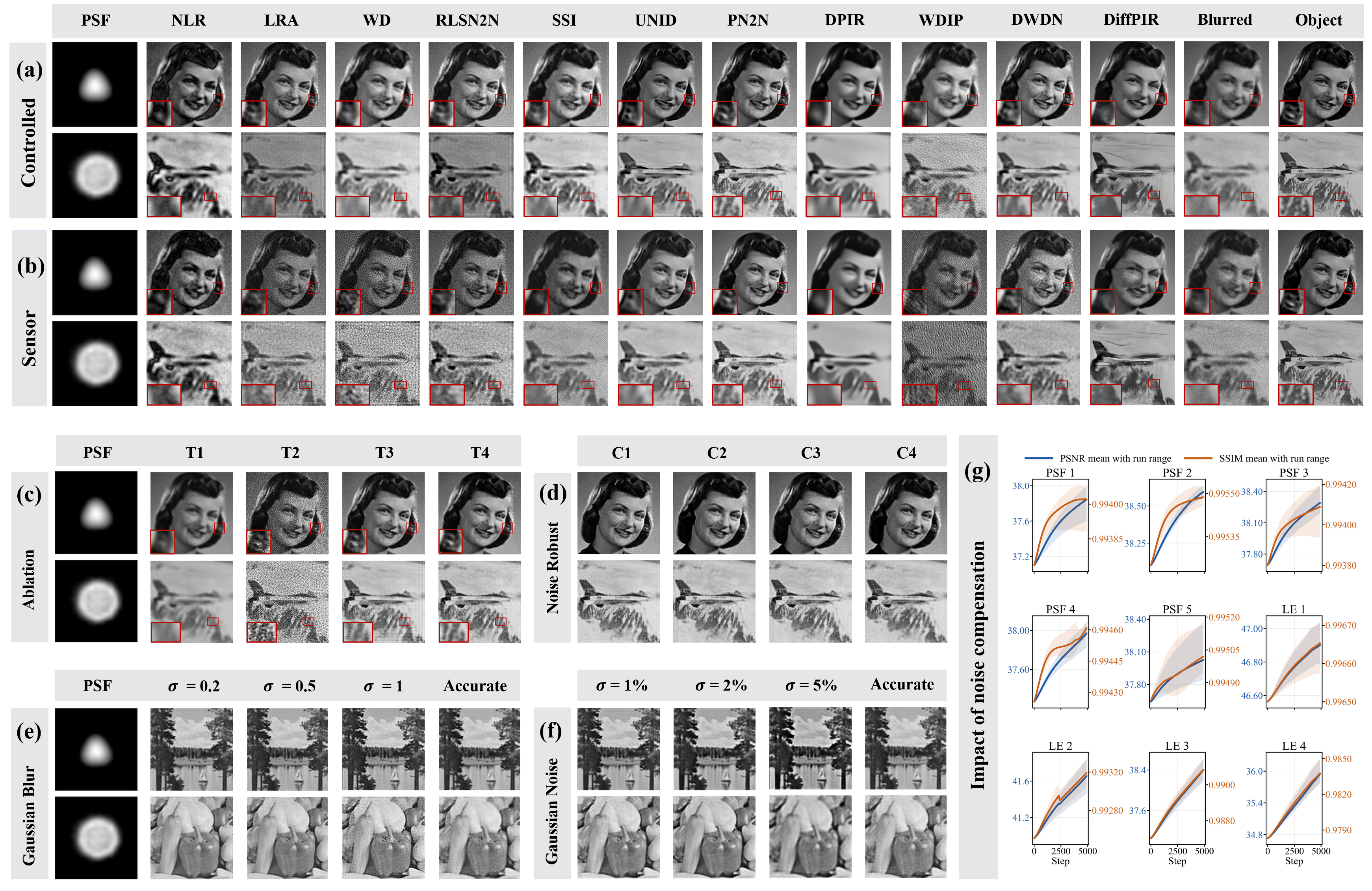}
    \caption{(a) Comparison of different methods on controlled biased noise datasets. (b) Comparison of different methods on multiple sensor simulation datasets. (c) Ablation study on controlled biased noise datasets. (d) Performance comparison of PN2N under various conditions. (e) Comparison of PN2N with and without PSF mismatch. (f) Comparison of PN2N with accurate and inaccurate PSF due to noise. (g) The impact of bias noise compensation on noisy blurred images, evaluated on both controlled biased noise datasets and multi-level sensor noise simulation datasets. The original images used are from \cite{deng2009imagenet, CVGUGR}.}
    \label{fig:simulation_results}
\end{figure*}

\subsection{Ablation study on controlled biased noise benchmark}
\label{subsec:ablation}

To isolate the effect of spatially varying biased noise and validate the theoretical analysis in Section \ref{sec:methods}, we first construct a controlled biased noise benchmark. Five PSFs corresponding to different apertures and defocus levels are measured in the laboratory and used to generate observations. For each PSF, ten representative images are selected from ImageNet \cite{deng2009imagenet} and CVG-UGR \cite{CVGUGR} datasets and degraded by convolution with the PSFs. The blurred images are then corrupted by a synthetic spatially varying biased-noise model:

\begin{equation}
n \sim \mathcal{P}(\mu_p) + \mathcal{N}(\mu_n, \sigma^2),
\label{eq:23}
\end{equation}
where $n$ denotes the per-pixel noise, $\mathcal{P}$ is a Poisson distribution, and $\mathcal{N}$ is a normal distribution. The parameters $\mu_n$, $\mu_p$, and $\sigma^2$ are pixel-dependent, determined by the spatial coordinates $(x, y)$ as defined in Eqs. (\ref{eq:24})–(\ref{eq:26}):

\begin{equation}
\mu_p = 0.001 \cdot \left( \dfrac{x}{4} \right)^2 + 0.02 \cdot y + 2,
\label{eq:24}
\end{equation}
\begin{equation}
\mu_n = 0.01 \cdot x + 0.01 \cdot y + 2,
\label{eq:25}
\end{equation}
\begin{equation}
\sigma^2 \sim U(20, 50),
\label{eq:26}
\end{equation}
where $U$ denotes a uniform distribution. Since the noise bias in real-world imaging remains constant over the entire acquisition session under stable environmental and thermal conditions \cite{wach2004noise}, we model it as shared across frames. In this controlled benchmark, $\mu_p$ and $\mu_n$ introduce spatially varying biased noise, while $\sigma^2$ controls the random Gaussian perturbation. This setting captures key simplified properties of spatially varying biased noise and is used for mechanism validation and ablation, whereas an additional multi-level sensor-noise benchmark is introduced in Section~\ref{subsec:sensor_noise}. We conducted ablation study on this benchmark, with results presented in Fig.~\ref{fig:simulation_results}(c), Table~\ref{tab:ablation1}, and Table~\ref{tab:ablation2}.

As can be seen from Fig.~\ref{fig:simulation_results}(c), Table~\ref{tab:ablation1}, and Table~\ref{tab:ablation2}, each module plays a critical role. The deblurring method with the bias-compensation term significantly outperforms variants without this term. This improvement can be attributed to the fact that the learnable bias term compensates for the stable expectation component of the biased noise in the blurred observation, thereby reducing the mismatch between the bias-corrected input and the physical blur-consistency model. This phenomenon validates the effectiveness of our proposed method and supports the theoretical analysis.

\noindent \textbf{Bias compensation analysis.} To verify that the learned bias term provides meaningful compensation rather than a trivial solution, we evaluate the bias-corrected observation during optimization. Since $b$ is used to correct the SN2N-initialized blurred observation before deblurring, we measure the similarity between $y'-b_t$ and the clean blurred image $Hx$, where $b_t$ denotes the learned bias at iteration $t$. Specifically, we report
\begin{equation}
\begin{split}
\mathrm{PSNR}_{\mathrm{corr}}(t) &= \mathrm{PSNR}(y' - b_t, PSF \otimes x), \\
\mathrm{SSIM}_{\mathrm{corr}}(t) &= \mathrm{SSIM}(y' - b_t, PSF \otimes x).
\end{split}
  \label{eq:27}
\end{equation}

Fig.~\ref{fig:simulation_results}(g) shows the convergence curves under the five PSFs. Although the SN2N already produces a high-quality blurred estimate, the corrected observation $y'-b_t$ consistently becomes closer to $Hx$ during optimization. This indicates that the learned bias term progressively compensates for the biased component rather than collapsing to arbitrary image content. 

\begin{table*}[htbp]
\centering
\caption{Configuration under different noise conditions in controlled biased noise simulation dataset.}
\label{tab:noise_condition}
\footnotesize
\setlength{\tabcolsep}{20pt} 
\renewcommand{\arraystretch}{0.8}
\begin{tabular}{@{}c c c c@{}}
\toprule
Condition & $\mu$ & $\sigma$ & Distribution \\
\midrule
C1 & $-$ & $\sigma^2 = 100$ & $\mathcal{N}(0, \sigma^2)$ \\
C2 & $\mu = 15$ & $-$ & $\mathcal{P}(\mu)$ \\
C3 & $\mu_p = 0.001 \cdot (x/4)^2 + 0.02 \cdot y + 2,\ \mu_n = 0.01 \cdot x + 0.01 \cdot y + 2$ & $\sigma^2 \sim U(20, 50)$ & $\mathcal{P}(\mu_p) + \mathcal{N}(\mu_n, \sigma^2)$ \\
C4 & $\mu_n = 0.001 \cdot (x/4)^2 + 0.02 \cdot y + 2,\ \mu_p = 0.01 \cdot x + 0.01 \cdot y + 2$ & $\sigma^2 \sim U(20, 50)$ & $\mathcal{P}(\mu_p) + \mathcal{N}(\mu_n, \sigma^2)$ \\
\bottomrule
\end{tabular}
\end{table*}

\begin{table*}[htbp]
\centering
\scriptsize
\caption{ Noise robustness comparison of PN2N and comparative methods.}
\label{tab:robust_condition}
\setlength{\tabcolsep}{0.4pt} 
\renewcommand{\arraystretch}{1}
\begin{tabular}{@{}c c *{22}{c} @{}} 
\toprule
& & \multicolumn{2}{c}{NLR} & \multicolumn{2}{c}{LRA} & \multicolumn{2}{c}{WD} & \multicolumn{2}{c}{SSI} & \multicolumn{2}{c}{UNID} & \multicolumn{2}{c}{RL-SN2N} & \multicolumn{2}{c}{PN2N} & \multicolumn{2}{c}{DPIR} & \multicolumn{2}{c}{WDIP} & \multicolumn{2}{c}{DiffPIR} & \multicolumn{2}{c}{DWDN}\\
& PSF & PSNR$\uparrow$ & SSIM$\uparrow$ & PSNR$\uparrow$ & SSIM$\uparrow$ & PSNR$\uparrow$ & SSIM$\uparrow$ & PSNR$\uparrow$ & SSIM$\uparrow$ & PSNR$\uparrow$ & SSIM$\uparrow$ & PSNR$\uparrow$ & SSIM$\uparrow$ & PSNR$\uparrow$ & SSIM$\uparrow$ & PSNR$\uparrow$ & SSIM$\uparrow$ & PSNR$\uparrow$ & SSIM$\uparrow$ & PSNR$\uparrow$ & SSIM$\uparrow$ & PSNR$\uparrow$ & SSIM$\uparrow$\\
\midrule
\multirow{2}{*}{\makecell{C1}}
& 3 & 16.52  & 0.6589 & 24.21 & 0.7904  & 27.10 & 0.8560 & 24.98 & 0.7675 & 24.02 & 0.8400 & 23.97 & 0.7575 & \textbf{27.69} & \textbf{0.8587} & 24.99 & 0.7698 & 22.49 & 0.5022 & 23.18 & 0.7016 & 26.63 & 0.8151\\
& 4 & 17.55  & 0.6176  &21.53 & 0.6902  & 24.09 & 0.7596 & 21.70 & 0.6685 & 22.71 & 0.7566 & 19.81 & 0.6799 & 25.50 & \textbf{0.7866} & 23.50 & 0.6889 & 18.87 & 0.2812 & 21.19 & 0.6169 & \textbf{25.67} & 0.7780\\
\midrule
\multirow{2}{*}{\makecell{C2}}
& 3 & 16.78 & 0.6916 & 25.93  & 0.8624  & 27.30 & 0.8692 & 25.06 & 0.8044 & 24.37 & 0.8652 & 24.39 & \textbf{0.8741} & \textbf{27.50} & 0.8562 & 24.95 & 0.7690 & 26.11 & 0.7270 & 23.16 & 0.7032 & 26.79 & 0.8695\\
& 4 & 17.78 & 0.6478  & 22.23 & 0.7642 & 24.19 & 0.7741 & 22.20 & 0.7549 & 22.83 & 0.7665 & 20.11 & 0.7352 & 26.15 & 0.7869 & 23.48 & 0.6875 & 23.06 & 0.5509 & 21.24 & 0.6192 & \textbf{26.70} & \textbf{0.8430}\\
\midrule
\multirow{2}{*}{\makecell{C3}}
& 3 & 16.82  & 0.6725  & 24.00  & 0.7788 & 26.29 & 0.8435 & 25.74 & 0.8553 & 24.84 & 0.8609 & 23.28 & 0.7850 & \textbf{28.08} & \textbf{0.8617} & 24.72 & 0.7678 & 21.78 & 0.4688 & 22.96 & 0.7011 & 26.19 & 0.7969\\
& 4 & 17.88  & 0.6261 & 20.82 & 0.6780 & 23.40 & 0.7463 & 22.07 & 0.6711 & 21.89 & 0.7170 & 19.59 & 0.6725 & \textbf{26.35} & \textbf{0.7872} & 23.44 & 0.6906 & 18.42 & 0.2556 & 21.13 & 0.6155 & 24.80 & 0.7643\\
\midrule
\multirow{2}{*}{\makecell{C4}}
& 3 & 16.73  & 0.6900 & 25.96 & 0.8823 & 27.30 & 0.8803 & 25.10 & 0.8134 & 24.57 & 0.8796 &  25.55 & 0.8798  & \textbf{27.52} & \textbf{0.8840} & 24.82 & 0.7682 & 21.78 & 0.4760 & 22.94 & 0.7003 & 26.32 & 0.7972\\
& 4 & 17.83  & 0.6296  & 21.58 & 0.6806 & 23.49 & 0.7465 & 22.16 & 0.6942 & 21.42 & 0.7015 & 20.03 & 0.6738 & \textbf{25.39} & \textbf{0.7790} & 23.43 & 0.6909 & 18.60 & 0.2621 & 21.15 & 0.6142 & 25.11 & 0.7641\\
\bottomrule
\end{tabular}
\end{table*}

\begin{table*}[htbp]
\centering
\small
\caption{Robustness of PN2N to PSF mismatch and measurement noise ($\lambda_2=0.01$).}
\label{tab:robust_PSF}
\setlength{\tabcolsep}{1.4pt} 
\footnotesize
\renewcommand{\arraystretch}{0.8}
\begin{tabular}{@{} c c c *{7}{c} @{}}
\toprule
\multirow{2}{*}{Regularization} 
& \multirow{2}{*}{PSF} 
& \multirow{2}{*}{Metric} 
& \multirow{2}{*}{Accurate} 
& \multicolumn{3}{c}{Gaussian blur (\(\sigma\))} 
& \multicolumn{3}{c}{Gaussian noise level (\(\sigma\))} \\
\cmidrule(lr){5-7} \cmidrule(lr){8-10}
& & & & 0.2 & 0.5 & 1 & 1\% & 2\% & 5\% \\
\midrule

\multirow{4}{*}{\makecell{$\times$}} 
& \multirow{2}{*}{3} 
& PSNR 
& $28.08 \pm 0.18$ & $28.32 \pm 0.49$ & $28.08 \pm 0.47$ & $26.89\pm0.48$  
& $27.53 \pm 0.47$ & $\textbf{26.03} \pm \textbf{0.17}$ & $21.61 \pm 0.18$ \\
& & SSIM 
& $0.8617 \pm 0.0142$ & $0.8128 \pm 0.0227$ & $0.8191 \pm 0.0045$ &  $0.7988\pm0.0069$ 
& $0.8259 \pm 0.0167$ & $\textbf{0.8167} \pm \textbf{0.0030}$ & $0.7558 \pm 0.0056$ \\

\cmidrule{2-10}

& \multirow{2}{*}{4} 
& PSNR 
& $26.35 \pm 0.47$ & $25.44 \pm 0.85$ & $\textbf{26.13} \pm \textbf{0.79}$ & $25.18 \pm 0.57$ 
& $25.46 \pm 0.40$ & $25.82 \pm 0.53$ & $23.86 \pm 0.15$ \\
& & SSIM 
& $0.7872 \pm 0.0167$ & $0.7126 \pm 0.0275$ & $\textbf{0.7316} \pm \textbf{0.0307}$ & $0.7010 \pm 0.0195$ 
& $0.7155 \pm 0.0151$ & $0.7442 \pm 0.0235$ & $0.7130 \pm 0.0105$ \\

\midrule

\multirow{4}{*}{\makecell{$\checkmark$}} 
& \multirow{2}{*}{3} 
& PSNR 
& $28.24 \pm 0.18$ & $28.28 \pm 0.17$ & $28.48 \pm 0.42$ & $27.05\pm0.28$ 
& $27.80 \pm 0.29$ & $\textbf{26.98} \pm \textbf{0.14}$ & $21.63 \pm 0.09$ \\
& & SSIM 
& $0.8274 \pm 0.0081$ & $0.8176 \pm 0.0079$ & $0.8254 \pm 0.0151$ & $ 0.7987\pm0.0080$
& $0.8310 \pm 0.0076$ & $\textbf{0.8199} \pm \textbf{0.0031}$ & $0.7587 \pm 0.0032$ \\

\cmidrule{2-10}

& \multirow{2}{*}{4} 
& PSNR 
& $25.44 \pm 0.56$ & $26.06 \pm 0.77$ & $\textbf{26.19} \pm \textbf{0.53}$ & $25.33 \pm 0.35$ 
& $25.54 \pm 0.43$ & $25.94 \pm 0.31$ & $24.25 \pm 0.28$ \\
& & SSIM 
& $0.7801 \pm 0.0173$ & $0.7282 \pm 0.0281$ & $\textbf{0.7370} \pm \textbf{0.0158}$ & $0.7160 \pm 0.0116$ 
& $0.7166 \pm 0.0091$ & $0.7498 \pm 0.0093$ & $0.7287 \pm 0.0147$ \\

\bottomrule
\end{tabular}
\end{table*}

\begin{table*}[htbp]
\centering
\small
\caption{Corrected-domain robustness of PN2N to PSF mismatch and measurement noise ($\lambda_2=0.01$).}
\label{tab:robust_PSF_corr}
\setlength{\tabcolsep}{0.7pt}
\footnotesize
\renewcommand{\arraystretch}{0.8}
\begin{tabular}{@{} c c c *{7}{c} @{}}
\toprule
\multirow{2}{*}{Regularization} 
& \multirow{2}{*}{PSF} 
& \multirow{2}{*}{Metric} 
& \multirow{2}{*}{Accurate} 
& \multicolumn{3}{c}{Gaussian blur (\(\sigma\))} 
& \multicolumn{3}{c}{Gaussian noise level (\(\sigma\))} \\
\cmidrule(lr){5-7} \cmidrule(lr){8-10}
& & & & 0.2 & 0.5 & 1 & 1\% & 2\% & 5\% \\
\midrule

\multirow{4}{*}{\makecell{$\times$}} 
& \multirow{2}{*}{3} 
& $\mathrm{PSNR}_{\mathrm{corr}}$ 
& $38.20 \pm 0.12$ & $38.31 \pm 0.25$ & $38.15 \pm 0.23$ & $38.17\pm0.10$ 
& $38.30 \pm 0.23$ & $\textbf{38.44} \pm \textbf{0.14}$ & $37.99 \pm 0.19$ \\
& & $\mathrm{SSIM}_{\mathrm{corr}}$ 
& $0.9940 \pm 0.0001$ & $0.9939 \pm 0.0002$ & $0.9937 \pm 0.0001$ & $ 0.9934\pm0.0003$ 
& $0.9932 \pm 0.0004$ & $\textbf{0.9922} \pm \textbf{0.0004}$ & $0.9895 \pm 0.0003$ \\

\cmidrule{2-10}

& \multirow{2}{*}{4} 
& $\mathrm{PSNR}_{\mathrm{corr}}$ 
& $37.75 \pm 0.22$ & $37.66 \pm 0.13$ & $\textbf{37.82} \pm \textbf{0.30}$ & $37.82 \pm 0.29$ 
& $37.81 \pm 0.21$ & $37.99 \pm 0.27$ & $38.02 \pm 0.24$ \\
& & $\mathrm{SSIM}_{\mathrm{corr}}$ 
& $0.9944 \pm 0.0009$ & $0.9944 \pm 0.0011$ & $\textbf{0.9944} \pm \textbf{0.0010}$ & $0.9943 \pm 0.0001$ 
& $0.9944 \pm 0.0001$ & $0.9943 \pm 0.0007$ & $0.9941 \pm 0.0022$ \\

\midrule

\multirow{4}{*}{\makecell{$\checkmark$}} 
& \multirow{2}{*}{3} 
& $\mathrm{PSNR}_{\mathrm{corr}}$ 
& $38.42 \pm 0.21$ & $38.24 \pm 0.21$ & $38.60 \pm 0.20$ & $38.52\pm0.10$
& $38.74 \pm 0.27$ & $\textbf{38.36} \pm \textbf{0.24}$ & $38.13 \pm 0.15$ \\
& & $\mathrm{SSIM}_{\mathrm{corr}}$ 
& $0.9941 \pm 0.0002$ & $0.9940 \pm 0.0001$ & $0.9940 \pm 0.0002$ & $0.9939\pm0.0001$
& $0.9938 \pm 0.0002$ & $\textbf{0.9931} \pm \textbf{0.0003}$ & $0.9905 \pm 0.0006$ \\

\cmidrule{2-10}

& \multirow{2}{*}{4} 
& $\mathrm{PSNR}_{\mathrm{corr}}$ 
& $37.76 \pm 0.23$ & $37.86 \pm 0.45$ & $\textbf{37.89} \pm \textbf{0.36}$ & $37.68 \pm 0.31$ 
& $37.68 \pm 0.06$ & $37.75 \pm 0.36$ & $37.95 \pm 0.29$ \\
& & $\mathrm{SSIM}_{\mathrm{corr}}$ 
& $0.9944 \pm 0.0024$ & $0.9943 \pm 0.0021$ & $\textbf{0.9944} \pm \textbf{0.0027}$ & $0.9942 \pm 0.0020$ 
& $0.9944 \pm 0.0001$ & $0.9943 \pm 0.0023$ & $0.9941 \pm 0.0002$ \\

\bottomrule
\end{tabular}
\end{table*}

\subsection{Stability analysis}
\label{subsec:robutst}

\noindent \textbf{Robustness to image noise.}
To further validate the robustness of PN2N, we conducted deblurring experiments under diverse noise conditions, as summarized in Table~\ref{tab:noise_condition}. The results are reported in Table~\ref{tab:robust_condition} and Fig.~\ref{fig:simulation_results}(d). PN2N remains robust across different noise distributions and PSF severities. Under the simple zero-mean Gaussian case C1, PN2N achieves the best SSIM for both PSFs, while its PSNR is only slightly lower than the supervised baseline DWDN on PSF 4 and remains the best on PSF 3. Under the pure Poisson case C2, several methods become competitive because the degradation better matches their assumed noise models; nevertheless, PN2N still remains among the top-performing methods. The advantage of PN2N becomes most evident under the mixed biased-noise cases C3 and C4, where both Poisson and Gaussian components have spatially varying nonzero expectations. In these settings, PN2N achieves the best PSNR and SSIM under both PSF 3 and PSF 4, showing that the proposed bias-adaptive formulation is particularly effective when the noise deviates from zero-mean or single-distribution assumptions. Traditional methods are sensitive to their assumed noise models, while recent robust methods such as UNID and SSI still lack an explicit mechanism for modeling stable biased noise. In contrast, PN2N compensates for the stable bias component while preserving the physical blur-consistency constraint, leading to more stable performance under complex biased noise.

\noindent \textbf{Robustness to PSF uncertainty.}
To evaluate robustness to PSF inaccuracies, we conducted simulated defocus deblurring experiments with two PSFs, where PSF 3 represents mild defocus and PSF 4 represents severe defocus. The PSF was corrupted by Gaussian blur to simulate defocus mismatch and by additive Gaussian noise to simulate PSF measurement error.

Table~\ref{tab:robust_PSF} reports the final deblurring results with mean and standard deviation. PN2N remains stable under small and moderate PSF perturbations. For PSF 3, mild PSF blur and low-level PSF noise only cause limited degradation, while severe corruption, especially 5\% PSF noise, leads to a clear performance drop. For PSF 4, the task is more sensitive because stronger defocus removes more high-frequency information, but PN2N still maintains acceptable performance under moderate mismatch and measurement noise. The bias regularizer does not universally increase final PSNR/SSIM under an accurate PSF, since the unregularized model can already fit the correct forward model. However, under mismatched or noisy PSFs, it generally improves stability and often improves final reconstruction, especially for severe blur and stronger perturbations, by preventing the bias term from absorbing PSF-mismatch residuals or scene-dependent structures.

Table~\ref{tab:robust_PSF_corr} further evaluates the corrected observation before deblurring. Compared with the final deblurring metrics in Table~\ref{tab:robust_PSF}, the corrected-domain metrics are much more stable. Under PSF mismatch, the regularizer stabilizes the bias compensation, and the corrected-domain metrics remain stable and often improve without suffering excessive degradation from PSF inaccuracies. Under PSF noise, bias compensation does not necessarily improve the corrected-domain metrics, since the overall blurring effect remains similar; nevertheless, the corrected-domain metrics stay stable, and the regularizer still yields improved deblurring performance. This comparison indicates that the learned bias compensation itself is robust to PSF perturbations, whereas the larger variation in final PSNR/SSIM mainly comes from the ill-posed deconvolution step under inaccurate PSFs.

\subsection{Quantitative evaluation with controlled simulation dataset}
\label{subsec:robust}

We compared the deblurring performance of various methods on the benchmark in Section \ref{subsec:ablation}. The deblurring results are shown in Fig. \ref{fig:simulation_results}(a) and Table \ref{tab:simulation_compare}. In quantitative comparisons under complex non-IID biased noise, PN2N achieves the best overall performance across the tested PSFs. Under mild defocus, the performance gap among methods is relatively small and UNID remains competitive. However, as defocus becomes stronger, high-frequency image details are increasingly attenuated, while biased noise introduces stronger perturbations in the same frequency range, making restoration substantially more challenging. Traditional methods degrade because their simplified noise assumptions are violated. SSI relies on mask-based denoising and struggles with spatially correlated biased noise, while RLSN2N lacks a mechanism for bias modeling. UNID improves robustness through Monte Carlo Dropout, but it tends to confuse fine structures with noise and may over-smooth details. DPIR is limited by the mismatch between its deep denoising prior and the biased degradation. WDIP, as a blind deconvolution baseline, becomes unreliable under stronger blur because complex noise interferes with PSF estimation. DWDN achieves competitive results, but its supervised training distribution does not fully match the tested biased-noise setting. DiffPIR benefits from a generative prior, yet under high noise it can introduce unnatural texture patterns, which degrades restoration fidelity. In contrast, PN2N explicitly compensates for the stable bias component while preserving PSF-based blur consistency, leading to more stable reconstruction under severe blur and complex biased noise.

\begin{figure*}[ht]
\centering\includegraphics[width=7in]{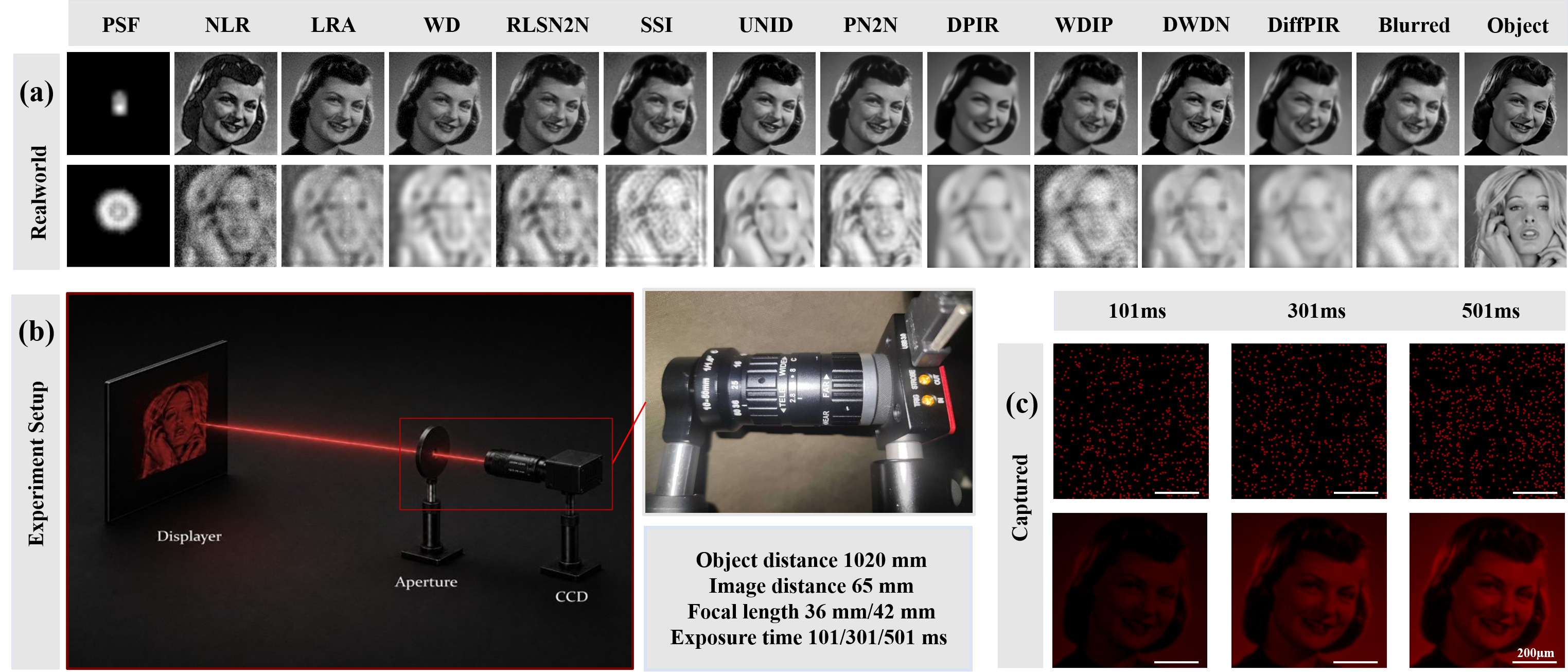}
    \caption{(a) Comparison of different methods on captured real data under Set 1 at 101 ms and Set 2 at 301 ms. (b) Experimental setup and equipment for real-world validation. (c) Captured low-light long-exposure images and dark-field noise captured without light signal. The original images are from \cite{deng2009imagenet, CVGUGR}.}
    \label{fig:real_results}
\end{figure*}

\subsection{Evaluation under multi-level sensor noise}
\label{subsec:sensor_noise}

To further evaluate the proposed method under low-light long-exposure degradation, we use a multi-level electron-domain sensor noise benchmark following the real complex noise composition of Pyxel~\cite{arko2022pyxel}. The noise parameters are specified level by level rather than derived from a particular calibrated sensor, producing noise levels from mild to severe long-exposure noise with progressively decreasing SNR.

\begin{table}[ht]
\centering
\setlength{\tabcolsep}{2pt} 
\caption{Parameters of the multi-level sensor noise benchmark.}
\label{tab:sensor_noise}
\footnotesize
\begin{tabular}{@{}lcccccccccc@{}}
\toprule
Level & \(\bar S\) & \(\bar D\) & \(\sigma_{read}\) & \(\sigma_{dcnu}\) & Tail ratio & \(k_\gamma\) & \(\theta_\gamma\) & \(\sigma_{row/col}\) & SNR & SNR(dB)\\
\midrule
LE1 & 150 & 5 & 2.0 & 0.3 & \(5\times10^{-6}\) & 2.0 & 10 & 0.2 & 11.9 & 21.5\\
LE2 & 115 & 8 & 3.0 & 0.6 & \(1\times10^{-5}\) & 2.0 & 16 & 0.3 & 10.0 & 20.0\\
LE3 & 90 & 10 & 3.5 & 0.8 & \(2\times10^{-5}\) & 2.0 & 22 & 0.4 & 8.5 & 18.6\\
LE4 & 70 & 16 & 5.5 & 1.2 & \(3\times10^{-5}\) & 2.0 & 28 & 1.0 & 6.5 & 16.2\\
\bottomrule
\end{tabular}
\end{table}

Given a clean defocused image \(B\), and convert it to the photoelectron expectation by
\begin{equation}
    S_l(x,y)=\alpha_l B(x,y),
  \label{eq:28}
\end{equation}
where \(l\) denotes the noise level. The conversion factor \(\alpha_l\) is fixed for each level, mimicking a fixed exposure setting. The reference signal level \(\bar S_l\) in Table~\ref{tab:sensor_noise} is used to set the electron-domain scale. The observation is simulated as
\begin{equation}
\begin{split}
Y_{e,l}(x,y)
&=
\operatorname{Poisson}(S_l(x,y))
+
\operatorname{Poisson}(D_l(x,y)) \\
&\quad
+
R_{row,l}(y)
+
R_{col,l}(x)
+
\eta_l(x,y),
\end{split}
  \label{eq:29}
\end{equation}
where \(\eta_l(x,y)\sim\mathcal N(0,\sigma_{read,l}^2)\). The first term is scene-dependent photon shot noise, the second term is dark-current shot noise, \(D_l\) is the stable dark-current expectation map, \(R_{row,l}\) and \(R_{col,l}\) are stable row/column readout patterns, and \(\eta_l\) is pixel-wise random read noise. The dark-current expectation is generated by a core-tail mixture:
\begin{equation}
D_l(x,y)
=
(1-M_l(x,y))D_{core,l}(x,y)
+
M_l(x,y)D_{tail,l}(x,y),
  \label{eq:30}
\end{equation}
where $D_{core,l}(x,y)=[\bar D_l+\Delta D_l(x,y)]_+$ with $\Delta D_l=\sigma_{dcnu,l}\operatorname{Norm}(W_l)$, \(W_l(x,y)\sim\mathcal N(0,1)\), \([\cdot]_+=\max(\cdot,0)\), and \(\operatorname{Norm}(\cdot)\) denoting zero-mean, unit-std normalization. The high-current tail is defined by \(M_l(x,y)\sim\operatorname{Bernoulli}(\rho_l)\) and \(D_{tail,l}(x,y)\sim\operatorname{Gamma}(k_{\gamma,l},\theta_{\gamma,l})\) with an optional level-dependent upper clipping to avoid isolated saturated hot pixels. The row and column patterns are generated as sparse banding artifacts. Specifically, 3\% of rows and 3\% of columns are randomly selected; their offsets are sampled from
\(\mathcal N(0,\sigma_{row,l}^2)\) and \(\mathcal N(0,\sigma_{col,l}^2)\), respectively assigned to the selected rows or columns.

For each noise level, \(D_l\), \(\Delta D_l\), \(M_l\), \(D_{tail,l}\), \(R_{row,l}\), and \(R_{col,l}\) are generated once and shared by all images and PSFs under that level. For each noisy realization, photon shot noise, dark-current shot noise, and random read noise are independently resampled. The final observation is obtained by normalizing \(Y_{e,l}\) with \(\alpha_l\), clipping it to \([0,1]\), and quantizing it to an 8-bit image. The parameters are summarized in Table~\ref{tab:sensor_noise}. \(\bar S_l\), \(\bar D_l\), \(\sigma_{read,l}\), \(\sigma_{dcnu,l}\), \(\rho_l\), \(k_{\gamma,l}\), \(\theta_{\gamma,l}\), and \(\sigma_{row/col,l}\) control the nominal signal level, mean dark current, read noise, DCNU strength, tail probability, Gamma tail distribution, and row/column fixed-pattern strength, respectively. These parameters are chosen to span mild-to-severe long-exposure degradation rather than to calibrate a specific camera sensor.

We use two representative measured PSFs, PSF~3 and PSF~4, corresponding to moderate and severe defocus blur, respectively. For each PSF and each noise level, the same set of ten images is used for all competing methods, and all methods are evaluated using identical noise realizations for fair comparison. As shown in Table~\ref{tab:le_results} and Fig~\ref{fig:simulation_results}(b), PN2N achieves the best performance under the multi-level sensor noise benchmark. Compared with the controlled biased noise benchmark, this setting introduces signal-dependent photon shot noise, stable dark-current bias, dark-current shot noise, dark-current non-uniformity, sparse high-current tail, row/column fixed-pattern components, and read noise. Although photon shot noise varies with scene content, the sensor-bias component is shared under the same acquisition condition. The results show that the proposed shared-bias formulation remains effective when scene-dependent photon noise and scene-invariant sensor bias coexist, especially under stronger long-exposure degradation.

For this benchmark, the reference stable bias is defined as
\begin{equation}
b_l^\star=\frac{D_l(x,y)+R_l^{row}(y)+R_l^{col}(x)}{\alpha_l},
  \label{eq:31}
\end{equation}
which excludes the scene-dependent photon shot noise and random zero-mean read noise. Fig.~\ref{fig:simulation_results}(g) shows the convergence of $y'-b_t$ toward $Hx$ under LE1--LE4 with PSF 4. The corrected observation becomes consistently closer to the clean blurred image across different noise levels.

\subsection{Experimental evaluation with captured images}
\label{subsec:realworld}

We evaluate the captured-image performance under six fixed configurations, consisting of two defocus levels and three exposure times. Set 1 corresponds to stronger defocus blur, while Set 2 contains comparatively clearer observations. For each defocus--exposure configuration, the optical setup and exposure time were fixed, as shown in Fig. \ref{fig:real_results}(b). Representative captures at different exposure times are shown in Fig.~\ref{fig:real_results}(c).

The 101 ms / Set 1 configuration follows the full pretrain--finetune capture protocol used in the original experiment. Specifically, we captured 10 target images selected from ImageNet~\cite{deng2009imagenet} and CVG-UGR~\cite{CVGUGR}; each target was recorded with 400 consecutive frames, yielding 4000 single images and 2000 noisy image pairs. These 10 targets were split into two non-overlapping groups: Set A contains 5 targets and 1000 image pairs for deblurring and evaluation, while Set B contains the remaining 5 targets and 1000 image pairs for SN2N pretraining. PN2N was evaluated on this configuration, and PF-PN2N used the SN2N model pretrained on Set B with only 100 image pairs from Set A for finetuning. For the other five configurations, we used the same fixed-setup acquisition procedure with the corresponding measured PSF and exposure time, but collected a compact evaluation set of 10 target images and 2000 captured frames per configuration for deblurring evaluation. These configurations were used to assess the robustness of PN2N and the competing baselines across different blur and exposure conditions.

The quantitative results are reported in Table~\ref{tab:defocus_results}, and visual comparisons are shown in Fig.~\ref{fig:real_results}(a). At 101 ms, the acquisition represents a weak-signal low-light condition with relatively limited dark-current accumulation. Although the overall sensor bias is not the strongest, the photon signal is weak, making noise-contaminated deblurring difficult. PN2N achieves the best SSIM on both defocus sets and the best PSNR on Set 2. On the more blurred Set 1, its PSNR is slightly lower than UNID, but its SSIM is higher, indicating better structural preservation. This is consistent with the visual results, where UNID tends to produce smoother outputs that may increase PSNR but introduce stronger structural distortion.

\begin{table*}[htbp]
\centering
\caption{Performance comparison under the multi-level sensor noise benchmark.}
\label{tab:le_results}
\scriptsize
\setlength{\tabcolsep}{0.55pt}
\renewcommand{\arraystretch}{1}
\begin{tabular}{@{}c c *{22}{c} @{}}
\toprule
& & \multicolumn{2}{c}{NLR} & \multicolumn{2}{c}{LRA} & \multicolumn{2}{c}{WD} & \multicolumn{2}{c}{SSI} & \multicolumn{2}{c}{UNID} & \multicolumn{2}{c}{RL-SN2N} & \multicolumn{2}{c}{PN2N} & \multicolumn{2}{c}{DPIR} & \multicolumn{2}{c}{WDIP} & \multicolumn{2}{c}{DiffPIR} & \multicolumn{2}{c}{DWDN}\\
& PSF & PSNR$\uparrow$ & SSIM$\uparrow$ & PSNR$\uparrow$ & SSIM$\uparrow$ & PSNR$\uparrow$ & SSIM$\uparrow$ & PSNR$\uparrow$ & SSIM$\uparrow$ & PSNR$\uparrow$ & SSIM$\uparrow$ & PSNR$\uparrow$ & SSIM$\uparrow$ & PSNR$\uparrow$ & SSIM$\uparrow$ & PSNR$\uparrow$ & SSIM$\uparrow$ & PSNR$\uparrow$ & SSIM$\uparrow$ & PSNR$\uparrow$ & SSIM$\uparrow$ & PSNR$\uparrow$ & SSIM$\uparrow$\\
\midrule
\multirow{2}{*}{\makecell{LE1}}
& 3 & 19.07 & 0.5688 & 27.63 & 0.6856 & 23.68 & 0.4685 & 24.90 & 0.6059 & 22.20 & 0.5929 & 27.38 & 0.7109 & \textbf{31.09} & \textbf{0.8746} & 26.88 & 0.7472 & 23.49 & 0.4548 & 25.49 & 0.6948 & 28.69 & 0.7586\\
& 4 & 19.55 & 0.5479 & 25.48 & 0.5991 & 21.75 & 0.3804 & 22.88 & 0.5273 & 21.84 & 0.5905 & 24.80 & 0.6173 & \textbf{29.22} & \textbf{0.7949} & 25.45 & 0.6981 & 20.17 & 0.2789 & 23.92 & 0.6297 & 27.98 & 0.7726\\
\midrule
\multirow{2}{*}{\makecell{LE2}}
& 3 & 19.33 & 0.5577 & 26.70 & 0.6390 & 22.58 & 0.4133 & 23.62 & 0.5844 & 25.89 & 0.7242 & 26.75 & 0.6699 & \textbf{31.58}& \textbf{0.8683} & 26.52 & 0.7416 & 21.19 & 0.3534 & 25.28 & 0.6915 & 27.94 & 0.7181\\
& 4 & 19.83 & 0.5361 & 24.89 & 0.5590 & 20.88 & 0.3341 & 22.43 & 0.4851 & 21.50 & 0.5520 & 24.57 & 0.5855 & \textbf{28.43} & \textbf{0.7552} & 25.17 & 0.6935 & 18.02 & 0.1867 & 23.80 & 0.6281 & 27.39 & 0.7465\\
\midrule
\multirow{2}{*}{\makecell{LE3}}
& 3 & 19.65 & 0.5455 & 25.74 & 0.5945 & 21.58 & 0.3657 & 23.09 & 0.5371 & 22.04 & 0.5778 & 26.16 & 0.6301 & \textbf{30.76} & \textbf{0.8424} & 26.05 & 0.7358 & 20.32 & 0.3152 & 24.96 & 0.6869 & 27.08 & 0.6779\\
& 4 & 20.16 & 0.5248 & 24.18 & 0.5193 & 20.00 & 0.2922 & 22.53 & 0.4831 & 22.05 & 0.5858 & 24.23 & 0.5517 & \textbf{28.23} & \textbf{0.7406} & 24.82 & 0.6887 & 17.82 & 0.1904 & 23.57 & 0.6244 & 26.57 & 0.7151\\
\midrule
\multirow{2}{*}{\makecell{LE4}}
& 3 & 20.55 & 0.5440 & 24.05 & 0.5359 & 20.25 & 0.3086 & 22.19 & 0.4964 & 20.89 & 0.5615 & 24.99 & 0.5692 & \textbf{29.88} & \textbf{0.8103} & 24.71 & 0.7204 & 17.93 & 0.2131 & 23.95 & 0.6772 & 25.38 & 0.6377\\
& 4 & 21.08 & 0.5168 & 22.93 & 0.4645 & 18.88 & 0.2431 & 20.88 & 0.4218 & 21.41 & 0.5586 & 23.50 & 0.4957 & \textbf{27.43} & \textbf{0.6970} & 23.76 & 0.6757 & 16.14 & 0.1344 & 22.84 & 0.6166 & 24.91 & 0.6702\\
\bottomrule
\end{tabular}
\end{table*}

\begin{table*}[htbp]
\centering
\caption{Comparison of the results of different methods on defocused experiment.}
\label{tab:defocus_results}
\scriptsize
\setlength{\tabcolsep}{0.55pt} 
\renewcommand{\arraystretch}{1}
\begin{tabular}{@{}c c *{22}{c} @{}} 
\toprule
& & \multicolumn{2}{c}{NLR} & \multicolumn{2}{c}{LRA} & \multicolumn{2}{c}{WD} & \multicolumn{2}{c}{SSI} & \multicolumn{2}{c}{UNID} & \multicolumn{2}{c}{RL-SN2N} & \multicolumn{2}{c}{PN2N} & \multicolumn{2}{c}{DPIR} & \multicolumn{2}{c}{WDIP} & \multicolumn{2}{c}{DiffPIR} & \multicolumn{2}{c}{DWDN}\\
& Set & PSNR$\uparrow$ & SSIM$\uparrow$ & PSNR$\uparrow$ & SSIM$\uparrow$ & PSNR$\uparrow$ & SSIM$\uparrow$ & PSNR$\uparrow$ & SSIM$\uparrow$ & PSNR$\uparrow$ & SSIM$\uparrow$ & PSNR$\uparrow$ & SSIM$\uparrow$ & PSNR$\uparrow$ & SSIM$\uparrow$ & PSNR$\uparrow$ & SSIM$\uparrow$ & PSNR$\uparrow$ & SSIM$\uparrow$ & PSNR$\uparrow$ & SSIM$\uparrow$ & PSNR$\uparrow$ & SSIM$\uparrow$\\
\midrule
\multirow{2}{*}{\makecell{101ms}}
& 1 & 15.13 & 0.2911 & 17.18 & 0.4444 & 17.66 & 0.4473 & 16.82 & 0.4662 & \textbf{18.85} & 0.5687 & 17.80 & 0.4930 & 18.78 & \textbf{0.5789} & 17.53 & 0.4841 & 16.53 & 0.3103 & 16.66 & 0.4314 & 17.72 & 0.5034\\
& 2 & 18.86 & 0.5230 & 23.24 & 0.5770 & 24.34 & 0.6309 & 23.94 & 0.7271 & 21.92 & 0.6261 & 24.62 & 0.7451 & \textbf{24.86} & \textbf{0.7611} & 24.17 & 0.6983 & 24.76 & 0.7527 & 24.02 & 0.7144 & 23.37 & 0.7224\\
\midrule
\multirow{2}{*}{\makecell{301ms}}
& 1 & 20.77 & 0.5654 & 23.39 & 0.6310 & 23.25 & 0.6277 & 22.15 & 0.5551 & 23.50 & 0.6328 & 23.45 & 0.6304 & \textbf{23.57} & \textbf{0.6483} & 22.75 & 0.6026 & 22.04 & 0.5386 & 21.79 & 0.5723 & 23.46 & 0.6394\\
& 2 & 19.96 & 0.6369 & 26.19 & 0.7469 & 26.59 & 0.7729 & 24.74 & 0.7662 & 24.88 & 0.7904 & 26.68 & 0.8121 & \textbf{26.76} & \textbf{0.8391} & 25.87 & 0.7546 & 26.61 & 0.8105 & 25.81 & 0.7647 & 26.28 & 0.8278\\
\midrule
\multirow{2}{*}{\makecell{501ms}}
& 1 & 21.04 & 0.5760 & 23.62 & 0.6437 & 23.19 & 0.6330 & 22.11 & 0.5688 & 23.58 & 0.6475 & 23.60 & 0.6405 & \textbf{23.69} & \textbf{0.6533} & 22.79 & 0.6107 & 22.35 & 0.5619 & 21.73 & 0.5748 & 23.55 & 0.6483\\
& 2 & 20.35 & 0.6724 & 26.49 & 0.7845 & 26.73 & 0.8004 & 25.42 & 0.7874 & 24.92 & 0.8095 & \textbf{26.81} & 0.8241 & \textbf{26.81} & 0.8349 & 26.08 & 0.7688 & 26.69 & 0.8179 & 25.96 & 0.7756 & 26.70 & \textbf{0.8409}\\
\bottomrule
\end{tabular}
\end{table*}

\begin{table}[ht]
\centering
\caption{Comparison of the results on pretrain-finetune paradigm. T stands for direct training, PF stands for pretrain-finetune paradigm, and Pretrain stands for pretraining only.}
\label{tab:PF}
\footnotesize
\setlength{\tabcolsep}{4.4pt}
\begin{tabular}{@{}l *{6}{c} @{}}
\toprule
 & Pretrain & T 100 & PF 100 & T 300 & PF 300 & T 1000 \\
\midrule
PSNR$\uparrow$ & 16.63 & 18.13 & \textbf{18.24} & 18.14 & \textbf{18.52} & \textbf{18.16} \\
SSIM$\uparrow$ & 0.3891 & 0.5488 & 0.5647 & 0.5718 & \textbf{0.5878} & \textbf{0.5875} \\
\bottomrule
\end{tabular}
\end{table}

\begin{table}[ht]
\hfill
\centering
\caption{Comparison of computational cost and model complexity }
\label{tab:runtime_step}
\footnotesize
\begin{tabular}{@{}lccccc@{}}
\toprule
Method & Params & MACs & Runtime & PSNR & SSIM \\
\midrule
NLR    & - & - & 2.0 s  & 17.88 dB  & 0.6261 \\
LRA    & - & - & 1.6 s  & 20.82 dB  & 0.6780 \\
WD     & - & - & 1 ms  & 23.40 dB  & 0.7463 \\
SSI    & 0.554 M & 1.291 G   & 10.8 min  & 22.07 dB  & 0.6711  \\
UNID   & 0.989 M & 11.230 G  & 3.5 min  & 21.89 dB  & 0.7170 \\
PN2N   & 8.696 M & 12.721 G  & 7.5 min  & \textbf{26.35 dB}  & \textbf{0.7872}\\
RL-SN2N   & 17.266 M  & 39.905 G  & 15 min  & 19.59 dB  & 0.6725\\

DWDN  & 7.047M   & 83.684G &   0.40 s &  24.80 dB&  0.7643 \\
WDIP & 3.592M   & 25.504G &  2.27 min  &  18.42 dB& 0.2556 \\
DPIR  & 32.639M  & 138.626G &  0.57 s &  23.44 dB&  0.6906 \\
DiffPIR  & 552.800M & 1118.927G &  4.02 s &  21.13 dB&  0.6155 \\
\bottomrule
\end{tabular}
\end{table}

At 301 ms, the photon signal becomes more sufficient, while long-exposure noise and stable sensor-dependent bias are also more pronounced. This setting better reflects the target low-light long-exposure degradation considered in this work. PN2N achieves the best PSNR and SSIM on both Set 1 and Set 2, demonstrating its ability to exploit the increased photon information while compensating for structured biased noise. At 501 ms, the accumulated photoelectrons further improve the effective signal quality, so several competing methods also benefit despite the stronger long-exposure noise. PN2N still achieves the best performance on the more blurred Set 1. On the clearer Set 2, PN2N ties for the best PSNR and is only slightly lower than the supervised DWDN baseline in SSIM. Overall, PN2N remains robust across weak-signal, balanced long-exposure, and high-photon regimes, with the clearest advantage appearing when defocus blur and sensor-dependent noise jointly degrade the observations. The PF-PN2N results in Table~\ref{tab:PF} show that the pretrain--finetune strategy can approach PN2N performance using substantially fewer image pairs.

\section{Discussion and limitation}
\label{sec:discussion}

The main contribution of PN2N is to enable self-supervised defocus deblurring under low-light long-exposure conditions, where noise is often biased and violates the commonly assumed zero-mean model. By introducing a learnable bias term and coupling it with the physical blur-consistency constraint, PN2N can exploit the residual mismatch caused by biased noise and progressively compensate for the noise component without requiring clean reference images.

The method still has several limitations. First, PN2N is a nonblind deblurring method and therefore requires a measured or estimated PSF. Although the experiments show robustness to moderate PSF mismatch and measurement noise, joint PSF estimation and deblurring remain important directions for future work. Second, PN2N assumes that the noise bias is approximately stable within a short acquisition sequence. This assumption is reasonable when exposure time, gain, temperature, and sensor operating conditions are fixed, since dark-current expectation and dark-signal non-uniformity are often stable over short periods. However, if the acquisition condition changes significantly, the bias statistics may drift and the learnable bias should be re-estimated.

Regarding computational efficiency, Table~\ref{tab:runtime_step} reports trainable parameters, single-forward MACs, runtime, and metrics, evaluated on the representative PSF4/C3 setting. Params and MACs measure backbone complexity under a unified input size, while runtime reflects the full restoration pipeline, including masking, sampling, iterative optimization, degradation-model operations, and implementation overhead. For deep-prior, plug-and-play, and diffusion-prior baselines, MACs indicate per-forward complexity rather than total inference cost, which also depends on the number of optimization or sampling steps. As a self-supervised optimization-based method, PN2N is slower than feed-forward supervised models, but it requires no paired training data and remains competitive among related self-supervised methods. Since SSI was evaluated on an RTX 2080 Ti and other methods on an RTX 3090 due to environment compatibility, runtime is reported as a practical reference rather than a hardware-normalized comparison. Future work will focus on blind extension and optimization acceleration.

\section{Conclusion and future work}
\label{sec:future}
We propose Physen-Noise2Noise, a novel self-supervised deep learning deblurring framework specifically designed for low-light, long-exposure, and defocused scenarios. This method incorporates frequency-domain constraints based on physical models and utilizes multi-frame data continuously captured within a short time interval. Through learnable noise bias, it integrates the initialization process of SN2N with the autoregressive learning mechanism of Bias-DeblurNet, ultimately achieving excellent performance in deblurring tasks under complex biased noise distributions. The framework effectively handles complex biased noise that is position-dependent, varies pixel-wise, and follows mixture distributions, under the least restrictive assumptions about noise characteristics. Its pretrain-finetune paradigm enhances model robustness and computational efficiency even with sparse training data. Future work will address limitations in PSF estimation accuracy and optimize computational speed for real-time processing.

\section*{Acknowledgement}
The authors acknowledge the support of National Natural Science Foundation of China (NSFC) under Grant Number 12571470, 12271526 and 62275077. The work was supported by the Major Scientific and Technological Innovation Platform Project of Hunan Province (2024JC1003) and High-level Talent Research Start-up Project Funding of Henan Academy of Sciences(Project N0.232019024). Part of this work was carried out in part using computing resources at the High Performance Computing Center of Central South University.

\bibliographystyle{IEEEtran}
\bibliography{IEEEexample}

@inproceedings{dong2021learning,
  title={Learning spatially-variant map models for non-blind image deblurring},
  author={Dong, Jiangxin and Roth, Stefan and Schiele, Bernt},
  booktitle={Proceedings of the IEEE/CVF conference on computer vision and pattern recognition},
  pages={4886--4895},
  year={2021}
}

@article{sanghvi2022photon,
  title={Photon limited non-blind deblurring using algorithm unrolling},
  author={Sanghvi, Yash and Gnanasambandam, Abhiram and Chan, Stanley H},
  journal={IEEE Transactions on Computational Imaging},
  volume={8},
  pages={851--864},
  year={2022},
  publisher={IEEE}
}

@article{zhang2023infwide,
  title={INFWIDE: Image and feature space Wiener deconvolution network for non-blind image deblurring in low-light conditions},
  author={Zhang, Zhihong and Cheng, Yuxiao and Suo, Jinli and Bian, Liheng and Dai, Qionghai},
  journal={IEEE Transactions on Image Processing},
  volume={32},
  pages={1390--1402},
  year={2023},
  publisher={IEEE}
}

@article{gong2020learning,
  title={Learning deep gradient descent optimization for image deconvolution},
  author={Gong, Dong and Zhang, Zhen and Shi, Qinfeng and Van Den Hengel, Anton and Shen, Chunhua and Zhang, Yanning},
  journal={IEEE transactions on neural networks and learning systems},
  volume={31},
  number={12},
  pages={5468--5482},
  year={2020},
  publisher={IEEE}
}

@article{wang2020phase,
  title={Phase imaging with an untrained neural network},
  author={Wang, Fei and Bian, Yaoming and Wang, Haichao and Lyu, Meng and Pedrini, Giancarlo and Osten, Wolfgang and Barbastathis, George and Situ, Guohai},
  journal={Light: Science \& Applications},
  volume={9},
  number={1},
  pages={77},
  year={2020},
  publisher={Nature Publishing Group UK London}
}

@article{hendriksen2020noise2inverse,
  title={Noise2inverse: Self-supervised deep convolutional denoising for tomography},
  author={Hendriksen, Allard Adriaan and Pelt, Dani{\"e}l Maria and Batenburg, K Joost},
  journal={IEEE Transactions on Computational Imaging},
  volume={6},
  pages={1320--1335},
  year={2020},
  publisher={IEEE}
}

@inproceedings{ren2020neural,
  title={Neural blind deconvolution using deep priors},
  author={Ren, Dongwei and Zhang, Kai and Wang, Qilong and Hu, Qinghua and Zuo, Wangmeng},
  booktitle={Proceedings of the IEEE/CVF conference on computer vision and pattern recognition},
  pages={3341--3350},
  year={2020}
}

@inproceedings{tang2023uncertainty,
  title={Uncertainty-aware unsupervised image deblurring with deep residual prior},
  author={Tang, Xiaole and Zhao, Xile and Liu, Jun and Wang, Jianli and Miao, Yuchun and Zeng, Tieyong},
  booktitle={Proceedings of the IEEE/CVF conference on computer vision and pattern recognition},
  pages={9883--9892},
  year={2023}
}

@article{chen2022nonblind,
  title={Nonblind image deconvolution via leveraging model uncertainty in an untrained deep neural network},
  author={Chen, Mingqin and Quan, Yuhui and Pang, Tongyao and Ji, Hui},
  journal={International Journal of Computer Vision},
  volume={130},
  number={7},
  pages={1770--1789},
  year={2022},
  publisher={Springer}
}

@article{kobayashi2020image,
  title={Image deconvolution via noise-tolerant self-supervised inversion},
  author={Kobayashi, Hirofumi and Solak, Ahmet Can and Batson, Joshua and Royer, Loic A},
  journal={arXiv preprint arXiv:2006.06156},
  year={2020}
}

@article{qin2025robust,
  title={Robust Unsupervised Deep Learning for Nonblind Image Deconvolution With Inaccurate Kernels},
  author={Qin, Xinran and Quan, Yuhui and Chen, Zhuojie and Ji, Hui},
  journal={IEEE Transactions on Neural Networks and Learning Systems},
  year={2025},
  publisher={IEEE}
}

@article{lim2020cyclegan,
  title={CycleGAN with a blur kernel for deconvolution microscopy: Optimal transport geometry},
  author={Lim, Sungjun and Park, Hyoungjun and Lee, Sang-Eun and Chang, Sunghoe and Sim, Byeongsu and Ye, Jong Chul},
  journal={IEEE Transactions on Computational Imaging},
  volume={6},
  pages={1127--1138},
  year={2020},
  publisher={IEEE}
}

@inproceedings{nair2022nbd,
  title={NBD-GAP: non-blind image deblurring without clean target images},
  author={Nair, Nithin Gopalakrishnan and Yasarla, Rajeev and Patel, Vishal M},
  booktitle={2022 IEEE international conference on image processing (ICIP)},
  pages={3431--3435},
  year={2022},
  organization={IEEE}
}

@inproceedings{ulyanov2018deep,
  title={Deep image prior},
  author={Ulyanov, Dmitry and Vedaldi, Andrea and Lempitsky, Victor},
  booktitle={Proceedings of the IEEE conference on computer vision and pattern recognition},
  pages={9446--9454},
  year={2018}
}

@inproceedings{wang2019image,
  title={Image deconvolution with deep image and kernel priors},
  author={Wang, Zhunxuan and Wang, Zipei and Li, Qiqi and Bilen, Hakan},
  booktitle={Proceedings of the IEEE/CVF International Conference on Computer Vision Workshops},
  pages={0--0},
  year={2019}
}

@inproceedings{chatterjee2011noise,
  title={Noise suppression in low-light images through joint denoising and demosaicing},
  author={Chatterjee, Priyam and Joshi, Neel and Kang, Sing Bing and Matsushita, Yasuyuki},
  booktitle={CVPR 2011},
  pages={321--328},
  year={2011},
  organization={IEEE}
}

@inproceedings{jin2020review,
  title={A review of an old dilemma: Demosaicking first, or denoising first?},
  author={Jin, Qiyu and Facciolo, Gabriele and Morel, Jean-Michel},
  booktitle={proceedings of the IEEE/CVF conference on computer vision and pattern recognition workshops},
  pages={514--515},
  year={2020}
}

@inproceedings{lee2022ap,
  title={Ap-bsn: Self-supervised denoising for real-world images via asymmetric pd and blind-spot network},
  author={Lee, Wooseok and Son, Sanghyun and Lee, Kyoung Mu},
  booktitle={Proceedings of the IEEE/CVF Conference on Computer Vision and Pattern Recognition},
  pages={17725--17734},
  year={2022}
}

@article{healey2002radiometric,
  title={Radiometric CCD camera calibration and noise estimation},
  author={Healey, Glenn E and Kondepudy, Raghava},
  journal={IEEE Transactions on Pattern Analysis and Machine Intelligence},
  volume={16},
  number={3},
  pages={267--276},
  year={2002},
  publisher={IEEE}
}

@article{zhang2017beyond,
  title={Beyond a gaussian denoiser: Residual learning of deep cnn for image denoising},
  author={Zhang, Kai and Zuo, Wangmeng and Chen, Yunjin and Meng, Deyu and Zhang, Lei},
  journal={IEEE transactions on image processing},
  volume={26},
  number={7},
  pages={3142--3155},
  year={2017},
  publisher={IEEE}
}

@inproceedings{liang2021swinir,
  title={Swinir: Image restoration using swin transformer},
  author={Liang, Jingyun and Cao, Jiezhang and Sun, Guolei and Zhang, Kai and Van Gool, Luc and Timofte, Radu},
  booktitle={Proceedings of the IEEE/CVF international conference on computer vision},
  pages={1833--1844},
  year={2021}
}

@inproceedings{chen2023masked,
  title={Masked image training for generalizable deep image denoising},
  author={Chen, Haoyu and Gu, Jinjin and Liu, Yihao and Magid, Salma Abdel and Dong, Chao and Wang, Qiong and Pfister, Hanspeter and Zhu, Lei},
  booktitle={Proceedings of the IEEE/CVF Conference on Computer Vision and Pattern Recognition},
  pages={1692--1703},
  year={2023}
}

@article{herbreteau2023normalization,
  title={Normalization-equivariant neural networks with application to image denoising},
  author={Herbreteau, S{\'e}bastien and Moebel, Emmanuel and Kervrann, Charles},
  journal={Advances in Neural Information Processing Systems},
  volume={36},
  pages={5706--5728},
  year={2023}
}

@inproceedings{li2024synthetic,
  title={From synthetic to real: A calibration-free pipeline for few-shot raw image denoising},
  author={Li, Ruoqi and Liu, Chang and Wang, Ziyi and Du, Yao and Yang, Jingjing and Bao, Long and Sun, Heng},
  booktitle={Proceedings of the IEEE/CVF Conference on Computer Vision and Pattern Recognition},
  pages={1106--1114},
  year={2024}
}

@article{lehtinen2018noise2noise,
  title={Noise2Noise: Learning image restoration without clean data},
  author={Lehtinen, Jaakko and Munkberg, Jacob and Hasselgren, Jon and Laine, Samuli and Karras, Tero and Aittala, Miika and Aila, Timo},
  journal={arXiv preprint arXiv:1803.04189},
  year={2018}
}

@inproceedings{krull2019noise2void,
  title={Noise2void-learning denoising from single noisy images},
  author={Krull, Alexander and Buchholz, Tim-Oliver and Jug, Florian},
  booktitle={Proceedings of the IEEE/CVF conference on computer vision and pattern recognition},
  pages={2129--2137},
  year={2019}
}

@inproceedings{huang2021neighbor2neighbor,
  title={Neighbor2neighbor: Self-supervised denoising from single noisy images},
  author={Huang, Tao and Li, Songjiang and Jia, Xu and Lu, Huchuan and Liu, Jianzhuang},
  booktitle={Proceedings of the IEEE/CVF conference on computer vision and pattern recognition},
  pages={14781--14790},
  year={2021}
}

@inproceedings{quan2020self2self,
  title={Self2self with dropout: Learning self-supervised denoising from single image},
  author={Quan, Yuhui and Chen, Mingqin and Pang, Tongyao and Ji, Hui},
  booktitle={Proceedings of the IEEE/CVF conference on computer vision and pattern recognition},
  pages={1890--1898},
  year={2020}
}

@article{qu2024self,
  title={Self-inspired learning for denoising live-cell super-resolution microscopy},
  author={Qu, Liying and Zhao, Shiqun and Huang, Yuanyuan and Ye, Xianxin and Wang, Kunhao and Liu, Yuzhen and Liu, Xianming and Mao, Heng and Hu, Guangwei and Chen, Wei and others},
  journal={Nature Methods},
  volume={21},
  number={10},
  pages={1895--1908},
  year={2024},
  publisher={Nature Publishing Group US New York}
}

@article{ma2025pixel2pixel,
  title={Pixel2Pixel: A Pixelwise Approach for Zero-Shot Single Image Denoising},
  author={Ma, Qing and Jiang, Junjun and Zhou, Xiong and Liang, Pengwei and Liu, Xianming and Ma, Jiayi},
  journal={IEEE Transactions on Pattern Analysis and Machine Intelligence},
  year={2025},
  publisher={IEEE}
}

@article{richardson1972bayesian,
  title={Bayesian-based iterative method of image restoration},
  author={Richardson, William Hadley},
  journal={Journal of the optical society of America},
  volume={62},
  number={1},
  pages={55--59},
  year={1972},
  publisher={Optical Society of America}
}

@article{lucy1974iterative,
  title={An iterative technique for the rectification of observed distributions},
  author={Lucy, Leon B},
  journal={Astronomical Journal, Vol. 79, p. 745 (1974)},
  volume={79},
  pages={745},
  year={1974}
}

@article{dhawan1985image,
  title={Image restoration by Wiener deconvolution in limited-view computed tomography},
  author={Dhawan, Atam Prakash and Rangayyan, Rangaraj M and Gordon, Richard},
  journal={Applied optics},
  volume={24},
  number={23},
  pages={4013--4020},
  year={1985},
  publisher={Optical Society of America}
}

@article{mukherjee2018imaging,
  title={Imaging through scattering medium by adaptive non-linear digital processing},
  author={Mukherjee, Saswata and Rosen, Joseph},
  journal={Scientific reports},
  volume={8},
  number={1},
  pages={10517},
  year={2018},
  publisher={Nature Publishing Group UK London}
}

@article{dong2020deep,
  title={Deep wiener deconvolution: Wiener meets deep learning for image deblurring},
  author={Dong, Jiangxin and Roth, Stefan and Schiele, Bernt},
  journal={Advances in Neural Information Processing Systems},
  volume={33},
  pages={1048--1059},
  year={2020}
}

@inproceedings{deng2009imagenet,
  title={Imagenet: A large-scale hierarchical image database},
  author={Deng, Jia and Dong, Wei and Socher, Richard and Li, Li-Jia and Li, Kai and Fei-Fei, Li},
  booktitle={2009 IEEE conference on computer vision and pattern recognition},
  pages={248--255},
  year={2009},
  organization={Ieee}
}

@book{goodman2005introduction,
  title={Introduction to Fourier optics},
  author={Goodman, Joseph W},
  year={2005},
  publisher={Roberts and Company publishers}
}

@article{hosseini2019convolutional,
  title={Convolutional deblurring for natural imaging},
  author={Hosseini, Mahdi S and Plataniotis, Konstantinos N},
  journal={IEEE Transactions on Image Processing},
  volume={29},
  pages={250--264},
  year={2019},
  publisher={IEEE}
}

@book{tian2000noise,
  title={Noise analysis in CMOS image sensors},
  author={Tian, Hui},
  year={2000},
  publisher={stanFord university}
}

@article{konnik2014high,
  title={High-level numerical simulations of noise in CCD and CMOS photosensors: review and tutorial},
  author={Konnik, Mikhail and Welsh, James},
  journal={arXiv preprint arXiv:1412.4031},
  year={2014}
}

@book{boreman2001modulation,
  title={Modulation transfer function in optical and electro-optical systems},
  author={Boreman, Glenn D},
  volume={4},
  year={2001},
  publisher={SPIE press Bellingham, Washington}
}

@inproceedings{monroy2025generalized,
  title={Generalized recorrupted-to-recorrupted: Self-supervised learning beyond gaussian noise},
  author={Monroy, Brayan and Bacca, Jorge and Tachella, Juli{\'a}n},
  booktitle={Proceedings of the Computer Vision and Pattern Recognition Conference},
  pages={28155--28164},
  year={2025}
}

@inproceedings{pang2021recorrupted,
  title={Recorrupted-to-recorrupted: Unsupervised deep learning for image denoising},
  author={Pang, Tongyao and Zheng, Huan and Quan, Yuhui and Ji, Hui},
  booktitle={Proceedings of the IEEE/CVF conference on computer vision and pattern recognition},
  pages={2043--2052},
  year={2021}
}

@inproceedings{gal2016dropout,
  title={Dropout as a bayesian approximation: Representing model uncertainty in deep learning},
  author={Gal, Yarin and Ghahramani, Zoubin},
  booktitle={international conference on machine learning},
  pages={1050--1059},
  year={2016},
  organization={PMLR}
}

@article{wiedemann2024deep,
  title={A deep learning method for simultaneous denoising and missing wedge reconstruction in cryogenic electron tomography},
  author={Wiedemann, Simon and Heckel, Reinhard},
  journal={Nature Communications},
  volume={15},
  number={1},
  pages={8255},
  year={2024},
  publisher={Nature Publishing Group UK London}
}

@article{takeda2007kernel,
  title={Kernel regression for image processing and reconstruction},
  author={Takeda, Hiroyuki and Farsiu, Sina and Milanfar, Peyman},
  journal={IEEE Transactions on image processing},
  volume={16},
  number={2},
  pages={349--366},
  year={2007},
  publisher={IEEE}
}

@article{zhang2005multiscale,
  title={Multiscale LMMSE-based image denoising with optimal wavelet selection},
  author={Zhang, Lei and Bao, Paul and Wu, Xiaolin},
  journal={IEEE Transactions on circuits and systems for video technology},
  volume={15},
  number={4},
  pages={469--481},
  year={2005},
  publisher={IEEE}
}

@article{dabov2007image,
  title={Image denoising by sparse 3-D transform-domain collaborative filtering},
  author={Dabov, Kostadin and Foi, Alessandro and Katkovnik, Vladimir and Egiazarian, Karen},
  journal={IEEE Transactions on image processing},
  volume={16},
  number={8},
  pages={2080--2095},
  year={2007},
  publisher={IEEE}
}

@article{beck2009fast,
  title={Fast gradient-based algorithms for constrained total variation image denoising and deblurring problems},
  author={Beck, Amir and Teboulle, Marc},
  journal={IEEE transactions on image processing},
  volume={18},
  number={11},
  pages={2419--2434},
  year={2009},
  publisher={IEEE}
}

@article{Wei:20,
author = {Xiao-Xiang Wei and Lei Zhang and Hua Huang},
journal = {Opt. Express},
number = {7},
pages = {10683--10704},
publisher = {Optica Publishing Group},
title = {High-quality blind defocus deblurring of multispectral images with optics and gradient prior},
volume = {28},
month = {Mar},
year = {2020},
doi = {10.1364/OE.390158},
}

@article{yang1996structure,
  title={Structure adaptive anisotropic image filtering},
  author={Yang, Guang-Zhong and Burger, Peter and Firmin, David N and Underwood, SR},
  journal={Image and Vision Computing},
  volume={14},
  number={2},
  pages={135--145},
  year={1996},
  publisher={Elsevier}
}

@article{Kageyama:20,
author = {Yuta Kageyama and Mariko Isogawa and Daisuke Iwai and Kosuke Sato},
journal = {Opt. Express},
number = {14},
pages = {20391--20403},
publisher = {Optica Publishing Group},
title = {ProDebNet: projector deblurring using a convolutional neural network},
volume = {28},
month = {Jul},
year = {2020},
doi = {10.1364/OE.396159},
}

@article{tian2020deep,
  title={Deep learning on image denoising: An overview},
  author={Tian, Chunwei and Fei, Lunke and Zheng, Wenxian and Xu, Yong and Zuo, Wangmeng and Lin, Chia-Wen},
  journal={Neural Networks},
  volume={131},
  pages={251--275},
  year={2020},
  publisher={Elsevier}
}

@article{izadi2023image,
  title={Image denoising in the deep learning era},
  author={Izadi, Saeed and Sutton, Darren and Hamarneh, Ghassan},
  journal={Artificial Intelligence Review},
  volume={56},
  number={7},
  pages={5929--5974},
  year={2023},
  publisher={Springer}
}

@article{elad2023image,
  title={Image denoising: The deep learning revolution and beyond—a survey paper},
  author={Elad, Michael and Kawar, Bahjat and Vaksman, Gregory},
  journal={SIAM Journal on Imaging Sciences},
  volume={16},
  number={3},
  pages={1594--1654},
  year={2023},
  publisher={SIAM}
}

@article{laine2019high,
  title={High-quality self-supervised deep image denoising},
  author={Laine, Samuli and Karras, Tero and Lehtinen, Jaakko and Aila, Timo},
  journal={Advances in neural information processing systems},
  volume={32},
  year={2019}
}

@inproceedings{wach2004noise,
  title={Noise modeling for design and simulation of color imaging systems},
  author={Wach, Hans B and Dowski, Edward R},
  booktitle={Color and Imaging Conference},
  volume={12},
  pages={211--216},
  year={2004},
  organization={Society of Imaging Science and Technology}
}

@article{tao2024legan,
  title={LEGAN: A low-light image enhancement generative adversarial network for industrial internet of smart-cameras},
  author={Tao, Jing and Wang, Junliang and Zhang, Peng and Zhang, Jie and Yung, Kai-Leung and Ip, Wai Hung},
  journal={Internet of Things},
  volume={25},
  pages={101054},
  year={2024},
  publisher={Elsevier}
}

@article{ai2020extreme,
  title={Extreme low-light image enhancement for surveillance cameras using attention U-Net},
  author={Ai, Sophy and Kwon, Jangwoo},
  journal={Sensors},
  volume={20},
  number={2},
  pages={495},
  year={2020},
  publisher={MDPI}
}

@misc{CVGUGR,
  author = {{Computer Vision Group, University of Granada}},
  title = {CVG-UGR Image Database},
  howpublished = {\url{https://ccia.ugr.es/cvg/dbimagenes/}},
  note = {Accessed Jan. 22, 2026}
}

@inproceedings{cao2023physics,
  title={Physics-guided iso-dependent sensor noise modeling for extreme low-light photography},
  author={Cao, Yue and Liu, Ming and Liu, Shuai and Wang, Xiaotao and Lei, Lei and Zuo, Wangmeng},
  booktitle={Proceedings of the IEEE/CVF Conference on Computer Vision and Pattern Recognition},
  pages={5744--5753},
  year={2023}
}

@article{el2005cmos,
  title={CMOS image sensors},
  author={El Gamal, Abbas and Eltoukhy, Helmy},
  journal={IEEE Circuits and Devices Magazine},
  volume={21},
  number={3},
  pages={6--20},
  year={2005},
  publisher={IEEE}
}

@ARTICLE{11175047,
  author={Yang, Hong and Yang, Xianqiang},
  journal={IEEE Transactions on Pattern Analysis and Machine Intelligence}, 
  title={Zero-Shot Learning for Limited Photon Budget Denoising in Structured Illumination Microscopy}, 
  year={2026},
  volume={48},
  number={2},
  pages={1309-1320},
  doi={10.1109/TPAMI.2025.3612886}}

@article{chen2024self,
  title={Self-supervised denoising for multimodal structured illumination microscopy enables long-term super-resolution live-cell imaging},
  author={Chen, Xingye and Qiao, Chang and Jiang, Tao and Liu, Jiahao and Meng, Quan and Zeng, Yunmin and Chen, Haoyu and Qiao, Hui and Li, Dong and Wu, Jiamin},
  journal={PhotoniX},
  volume={5},
  number={1},
  pages={4},
  year={2024},
  publisher={Springer}
}

@article{qiao2024zero,
  title={Zero-shot learning enables instant denoising and super-resolution in optical fluorescence microscopy},
  author={Qiao, Chang and Zeng, Yunmin and Meng, Quan and Chen, Xingye and Chen, Haoyu and Jiang, Tao and Wei, Rongfei and Guo, Jiabao and Fu, Wenfeng and Lu, Huaide and others},
  journal={Nature communications},
  volume={15},
  number={1},
  pages={4180},
  year={2024},
  publisher={Nature Publishing Group UK London}
}

@article{zhang2021plug,
  title={Plug-and-play image restoration with deep denoiser prior},
  author={Zhang, Kai and Li, Yawei and Zuo, Wangmeng and Zhang, Lei and Van Gool, Luc and Timofte, Radu},
  journal={IEEE Transactions on Pattern Analysis and Machine Intelligence},
  volume={44},
  number={10},
  pages={6360--6376},
  year={2021},
  publisher={IEEE}
}

@inproceedings{bredell2023wiener,
  title={Wiener guided dip for unsupervised blind image deconvolution},
  author={Bredell, Gustav and Erdil, Ertunc and Weber, Bruno and Konukoglu, Ender},
  booktitle={Proceedings of the IEEE/CVF Winter Conference on Applications of Computer Vision},
  pages={3047--3056},
  year={2023}
}

@inproceedings{zhu2023denoising,
  title={Denoising diffusion models for plug-and-play image restoration},
  author={Zhu, Yuanzhi and Zhang, Kai and Liang, Jingyun and Cao, Jiezhang and Wen, Bihan and Timofte, Radu and Van Gool, Luc},
  booktitle={Proceedings of the IEEE/CVF conference on computer vision and pattern recognition},
  pages={1219--1229},
  year={2023}
}

@article{arko2022pyxel,
  title={Pyxel 1.0: an open source Python framework for detector and end-to-end instrument simulation},
  author={Arko, Matej and Prod’homme, Thibaut and Lemmel, Fr{\'e}d{\'e}ric and Serra, Benoit and George, Elizabeth and Kelman, Bradley and Pichon, Thibault and Biancalani, Enrico and Gilbert, James},
  journal={Journal of Astronomical Telescopes, Instruments, and Systems},
  volume={8},
  number={4},
  pages={048002--048002},
  year={2022},
  publisher={Society of Photo-Optical Instrumentation Engineers}
}

\vfill

\end{document}